\documentclass[10pt,twocolumn,letterpaper]{article}

\usepackage{cvpr}      %
\usepackage{graphicx}

\usepackage{amsmath,amsfonts,bm}

\def\eqref#1{equation~\ref{#1}}

\def\1{\bm{1}}

\DeclareMathAlphabet{\mathsfit}{\encodingdefault}{\sfdefault}{m}{sl}
\SetMathAlphabet{\mathsfit}{bold}{\encodingdefault}{\sfdefault}{bx}{n}

\usepackage{url}
\usepackage{booktabs}

\usepackage{nicefrac}
\usepackage{arydshln}
\usepackage{array}
\usepackage{adjustbox}
\usepackage{array, makecell} 
\usepackage{tabularx}   
\usepackage{booktabs}   

\usepackage{subcaption} %

\usepackage{color, colortbl}
\usepackage{pifont} %
\definecolor{Gray}{gray}{0.9}
\definecolor{brightturquoise}{rgb}{0.85, 1, 1}
\definecolor{newpurple}{HTML}{BC61F5}%
\newcommand{\mycc}{\cellcolor[HTML]{EDF6FF}}

\usepackage{pifont}
\newcommand{\cmark}{\ding{51}}  %
\newcommand{\xmark}{\ding{55}}  %
\definecolor{cvprblue}{rgb}{0.21,0.49,0.74}
\usepackage[pagebackref,breaklinks,colorlinks,citecolor=cvprblue]{hyperref}

\def\paperID{20429} %
\def\confName{CVPR}
\def\confYear{2026}
\usepackage{multirow}

\newcommand{\reffig}[1]{\text{Figure~\ref{#1}}}
\newcommand{\reftab}[1]{\text{Table~\ref{#1}}}

\newcommand\blfootnote[1]{%
  \begingroup
  \renewcommand\thefootnote{}\footnote{#1}%
  \addtocounter{footnote}{-1}%
  \endgroup
}

\title{VisRes Bench: On Evaluating the Visual Reasoning Capabilities of VLMs}

\author{
Brigitta Malagurski Törtei$^{1}$ \textsuperscript{*} \hspace{1em} Yasser Dahou$^{1}$ \textsuperscript{*}  \hspace{2em} Ngoc Dung Huynh$^{1}$ \textsuperscript{*}  \\ \hspace{2em} Wamiq Reyaz Para$^{1}$ \hspace{1em} Phúc H. Lê Khac$^{1}$\hspace{2em}  Ankit Singh$^{1}$\hspace{2em} \\   Sofian Chaybouti$^{1,2}$\hspace{2em} Sanath Narayan$^{1}$ \vspace{2mm} \\
$^{1}$Technology Innovation Institute, Abu Dhabi, UAE \\
$^{2}$Tuebingen AI Center/University of Tuebingen 
}

\begin{document}

\maketitle

\begin{abstract}

Vision-Language Models (VLMs) have achieved remarkable progress across tasks such as visual question answering and image captioning. Yet, the extent to which these models perform visual reasoning as opposed to relying on linguistic priors remains unclear. To address this, we introduce VisRes Bench, a benchmark designed to study visual reasoning in naturalistic settings without contextual language supervision. Analyzing model behavior across three levels of complexity, we uncover clear limitations in perceptual and relational visual reasoning capacities. VisRes isolates distinct reasoning abilities across its levels. Level 1 probes perceptual completion and global image matching under perturbations such as blur, texture changes, occlusion, and rotation; Level 2 tests rule-based inference over a single attribute (e.g., color, count, orientation); and Level 3 targets compositional reasoning that requires integrating multiple visual attributes. Across more than 19,000 controlled task images, we find that state-of-the-art VLMs perform near random under subtle perceptual perturbations, revealing limited abstraction beyond pattern recognition. We conclude by discussing how VisRes provides a unified framework for advancing abstract visual reasoning in multimodal research.
\blfootnote{\textsuperscript{*} Equal contribution}

\end{abstract}

\section{Introduction}

\begin{figure*}[t]
    \centering
    \includegraphics[width=\textwidth]{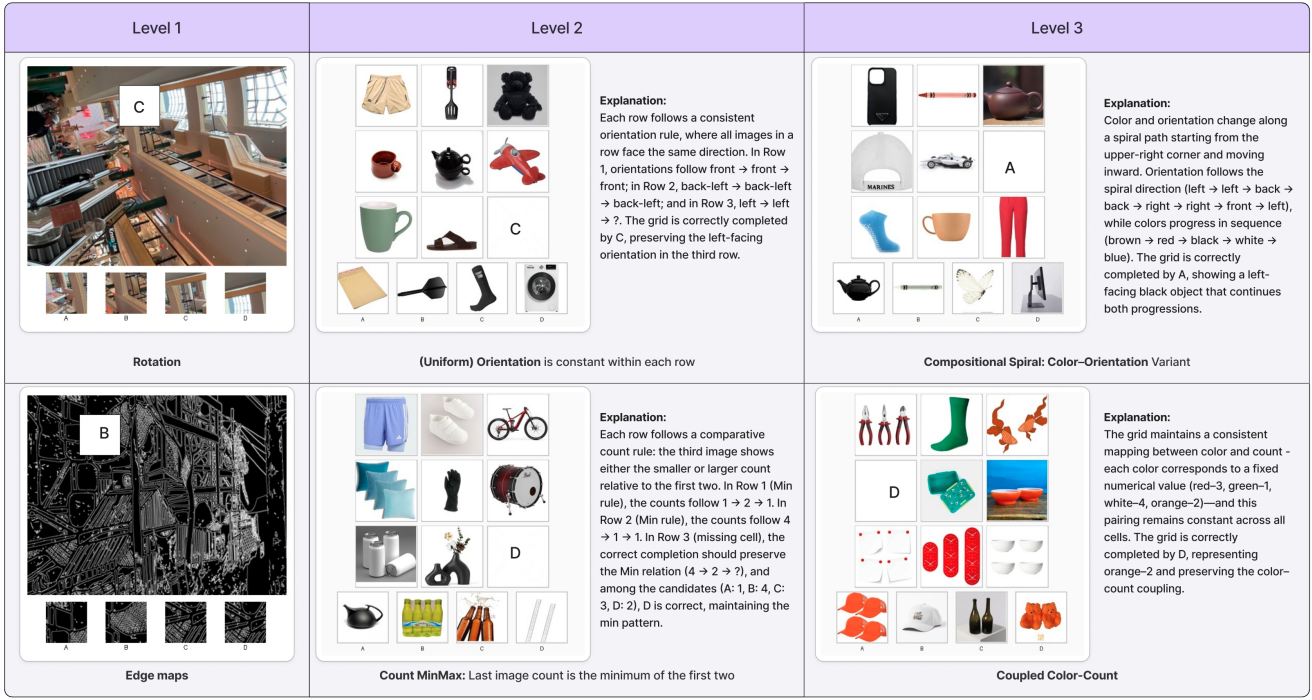}
    \caption{\textbf{Real samples from each level.} Level~1 (top) involves direct visual completion and matching without explicit rule inference (e.g., patch-C correctly continues the ceiling texture compared to patch-D), while Levels~2 and~3 (bottom) require increasingly complex rule-based reasoning over perceptual attributes. Accurate perception of individual attributes is necessary but not sufficient for solving compositional tasks. Current VLMs show poor performance on these compositional tasks. See Section~\ref{sec:performance}.}
    
    \label{fig:visres}
\end{figure*}

Humans reliably complete partial visual information, reconstructing occluded objects, continuing interrupted patterns, and inferring abstract rules from spatial arrangements. This ability draws on multiple levels of visual processing—from perceptual organization via Gestalt principles~\cite{Wagemans2012} and amodal completion~\cite{Morgan2016CorticalFeedback, Thielen2019AmodalCompletion} to attribute recognition and compositional reasoning—supporting stable object identity, spatial continuity, and causal understanding.

In contrast, VLMs, despite excelling at captioning and visual question answering, fail at these fundamental tasks when linguistic guidance is removed, unable to reconstruct partially visible components or maintain spatial continuity~\cite{Liu2025BeyondVisible}. These perceptual deficits cascade into higher-level reasoning failures~\cite{Zhou2025}. Without textual input, VLMs struggle with compositional reasoning and even simple attribute-based relations, with performance collapsing on multi-attribute tasks~\cite{Ke2025}. Text-conditioned competence does not transfer to purely visual benchmarks, revealing that apparent reasoning capabilities may reflect linguistic priors rather than genuine visual understanding.

Findings from cognitive neuroscience indicate that relational reasoning develops along a graded perceptual–conceptual continuum, where early visual representations support increasingly abstract relational and compositional inference~\cite{Holyoak2021Emergence, Bai2025CorePerception}. This continuum progresses hierarchically: perceptual grounding (recovering object properties from visual input) supports single-attribute reasoning (e.g., tracking color or count transformations), which in turn enables multi-attribute composition (e.g., reasoning jointly over color, count, and spatial relations). 
Because these stages build on one another, failures in perceptual extraction propagate upward: models that cannot form reliable visual representations cannot perform rule-based or compositional reasoning over them. Current VLMs exhibit fragile generalization under perceptual variation, highlighting the need for benchmarks that jointly evaluate perceptual and reasoning stages and diagnose where visual processing breaks down ~\cite{Lake2017Building, ilievski2024aligning}.

We introduce \textbf{VisRes}, a benchmark that spans this progression from perceptual grounding to structured visual reasoning in real-world imagery. The three levels of VisRes operationalize increasing reasoning demands: perceptual reconstruction (recovering structure under occlusion and degradation), single-attribute rule abstraction (inference over single visual attributes in Raven-style grids~\cite{raven1983manual}), and multi-attribute composition (integrating several concurrently varying attributes whose rules may be independent, coupled, or spatially organized). This setting defines task complexity in terms of the type of reasoning required—ranging from perceptual completion to rule abstraction to compositional integration—and enables principled diagnosis of where model failures arise. Our image-only, multiple-choice format minimizes linguistic priors, ensuring that performance reflects visual reasoning rather than textual shortcuts. Our key contributions are: 

\begin{itemize}
    \item VisRes structures task complexity across perceptual, single-attribute, and multi-attribute reasoning, providing a systematic framework for diagnosing visual reasoning capabilities.

    \item We systematically evaluate leading vision–language models and uncover distinct failure modes across levels- from spatial completion deficits to incomplete rule extraction to breakdowns in multi-attribute reasoning. 

    \item We demonstrate that VLMs' apparent reasoning capabilities collapse without textual context, exposing fundamental limitations in visual grounding and highlighting the need for architectures that integrate perception and abstraction more effectively.
\end{itemize}

\section{Related Work}

LLM benchmarks have seen a recent focus on reasoning through works such as ARC-AGI \cite{chollet2025arc}, AIME, Enigmata \cite{chen2025enigmata}, BBH \cite{suzgun2023challenging}, and Kor-bench \cite{ma2024kor}. These assess an LLM’s capacity to perform systematic, compositional, and multi-step reasoning, typically evaluating abilities such as abstraction, logical deduction, mathematical problem solving, and generalization under distribution shifts. However, they do not assess visual–spatial reasoning, a fundamental aspect of intelligence. Vision-language benchmarks initially focused on recognition, captioning, and VQA \cite{Kim2025, Pantazopoulos2025}, but recent work has begun to study structured reasoning, notably revealing that VLMs’ performance substantially declines without linguistic guidance \cite{AlTahan2024UniBench, vo2025visionlanguagemodelsbiased}.

This progression has exposed critical gaps in visual reasoning. Occlusion and amodal completion datasets \cite{Ao2023AmodalCompletionSurvey} and benchmarks like BLINK \cite{Fu2024Blink}, which assess a broad range of core perceptual skills, reveal persistent limitations. SalBench \cite{dahou2025vision0language} further shows that VLMs struggle with simple saliency-based odd-one-out detection, indicating deficits in early visual saliency and low-level feature discrimination. Controlled synthetic environments such as CLEVR \cite{Johnson2017CLEVR}, CLEVRER \cite{Yi2019CLEVRER}, PGM \cite{Barrett2018}, RAVEN \cite{Zhang2019RAVEN}, MARVEL \cite{Jiang2024MARVEL}, ARC-AGI-2 \cite{Chollet2025}, and CVR \cite{Zerroug2022CVR} have been instrumental for studying visual reasoning under tightly controlled settings, but have limited transfer to real-world imagery. More naturalistic reasoning tasks—including Bongard-OpenWorld \cite{wu2023bongardopenworld}, Bongard-HOI \cite{Jiang2022BongardHOI}, and V-PROM \cite{Teney2020V-PROM} have advanced relational reasoning in realistic imagery and helped motivate our design, but typically operate within a single domain and lack level-structured tasks. 
VisuLogic \cite{Xu2025VisuLogic} and VERIFY \cite{bi2025verify} focus on synthetic, diagram-based quantitative and spatial reasoning, while VISFACTOR \cite{Huang2025VisFactor} probes fundamental visual cognition using psychometric-inspired subtests. These benchmarks extend reasoning evaluation in controlled settings, but typically rely on synthetic stimuli and do not structure tasks across perceptual, single-attribute, and multi-attribute level.

Recent neuroscience-inspired frameworks separate perceptual and reasoning processes \cite{vaishnav2025cognitive}, aligning with evidence that relational inference depends on earlier perceptual organization \cite{Kuhnke2023HierarchicalConceptual, Holyoak2021Emergence, frontiers2014perception}. Yet current benchmarks seldom cover this full perceptual–reasoning continuum. Surveys highlight the need for evaluations that tightly couple perception with inference while minimizing linguistic cues \cite{Ke2025, Zhou2025, Su2025ThinkingWithImages}. VisRes addresses this gap by offering natural-image tasks that systematically vary perceptual demands and reasoning complexity.

\section{VisRes Overview}

\begin{figure}[t]
    \centering
    \includegraphics[width=\columnwidth]{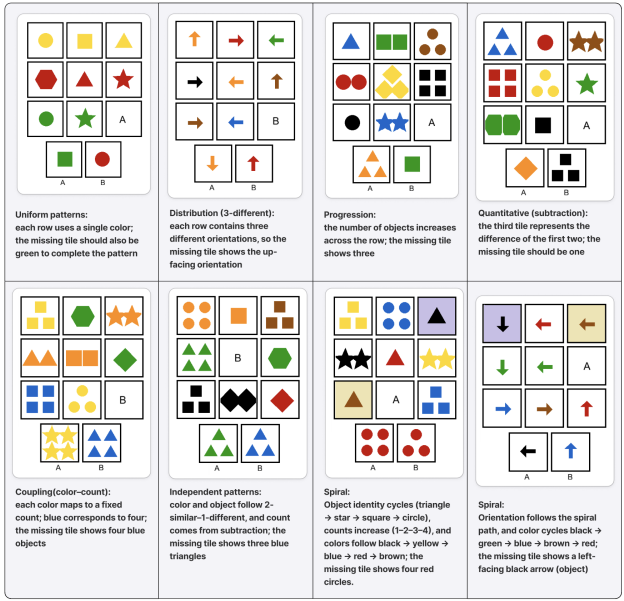}
    \caption{\textbf{Illustrative pattern rules used in Levels 2 and 3 tasks}. \textbf{Top:} Level-2 tasks where one attribute varies across the row. \textbf{Bottom:} Level-3 tasks where multiple attributes vary. See Section~\ref{sec:task_gen}.}
    \label{fig:visres_explainer}
\end{figure}

\begin{table*}[!htbp] 
    \centering
    \setlength{\tabcolsep}{10pt}
 \adjustbox{width=0.95\textwidth}{
\begin{tabular}
{l l c c c c c c c c c c c c} 
 \toprule[0.1em]

   & \textbf{Setting} 
 & \rotatebox{70}{\textbf{GPT-5}}  
 & \rotatebox{70}{\textbf{GPT-4o} }  
 & \rotatebox{70}{\textbf{Gemini-2.5} } 
 & \rotatebox{70}{\textbf{Qwen3-VL-4B }}  
 & \rotatebox{70}{\textbf{Qwen3-VL-8B }}  
 & \rotatebox{70}{\textbf{Qwen3-VL-30B }}  
 & \rotatebox{70}{\textbf{{Qwen3-VL-32B}}}
 & \rotatebox{70}{\textbf{Intern3.5-VL-4B}} 
 & \rotatebox{70}{\textbf{Intern3.5-VL-8B} } 
 & \rotatebox{70}{\textbf{MiMo-VL-7B-RL}} 
 & \rotatebox{70}{\textbf{GLM-4.1V-9B}} 
 & \rotatebox{70}{\textbf{Kimi-VL-A3B}} \\
 \toprule[0.1em]
 
\multirow{6}{*}{\textbf{Level-1}}
& Edges                  &   \textbf{27.17}             & 23.91  & 25.00 & 16.67 & 24.18 &  25.00 & 20.22 & 25.70 & 23.70  & 22.30 & 23.91 & 23.08 \\
& Location               &   23.71             & 20.62  & 26.00 & 23.16 & 22.73 &  22.40 & \textbf{27.27} & 25.10 & 22.30  & 25.77 & 25.41 & 18.09 \\
& Rotation               &   35.42    & 26.04  & 34.38 & \textbf{37.50}  & 34.04 &  36.05 & 33.33  & 25.70 & 25.90  & 29.17 & 29.17 & 24.20\\
& Brightness             &   25.26             & 27.37  & 27.37 &  \textbf{31.52} & 29.67 &  29.47 & 27.59 & 24.6& 22.20  & 27.37 & 27.38 & 28.57 \\

& Blur                   &   \textbf{31.18}    & 25.26  & 26.32 &  24.73 & 25.11 &  24.28 & 24.21 & 25.00 & 23.00  & 26.32 & 24.39 & 30.20\\
& Global@50\%            &   42.86             & 20.88  & \textbf{57.14} &  37.50 & 40.45 & 47.25 &  50.55 & 25.00  & 35.50  & 48.35 & 24.18 & 32.61\\
& Global@80\%            &   32.61             & 22.83  & \textbf{36.96}  &  25.88 & 27.27  &  35.87& 30.43 & 25.00 &  25.80  & 30.43 & 29.35 & 27.78\\
& \mycc \textbf{Average} &  \mycc  31.10    &  \mycc 23.86  &  \mycc \textbf{33.28} &  \mycc 28.17 &  \mycc 29.06 &  \mycc 31.20 &  \mycc 30.56 &  \mycc 25.16 &  \mycc 25.49 & \mycc 29.22 &  \mycc 26.26 &  \mycc 26.36\\
\midrule
\multirow{11}{*}{\textbf{Level-2}} 
25.& Uniform Orientation                  & 22.22  & 25.25  &  26.53 & 26.00 & 22.18 & 23.00 & 22.27 & \textbf{30.13}  & 21.18 & 19.19 & 22.10 & 20.20 \\
& Uniform Color                        & 96.00  & 21.00  & \textbf{97.00} & 66.20 & 83.67 & 88.00 & 91.43 & 27.80  & 26.60 & 78.95 & 71.72 & 45.95 \\
& Uniform Count                        & 61.00  & 25.00  & \textbf{90.91} & 40.82 & 56.36 & 59.00 & 60.10 & 27.80  & 29.20 & 52.75 & 46.24 & 51.48 \\
& 3-different Orientation              & 34.69  & 24.00  & 31.96 & 21.47 & 22.22 & 17.00 & 22.73 & 27.20  & 28.40 & 24.24 & \textbf{36.90} & 19.00 \\
& 3-different Color                    & \textbf{88.00}  & 26.00  & 69.00 & 45.92 & 51.82 & 65.00 & 63.64 & 22.80  & 25.80 & 46.46 & 38.95 & 39.47 \\
& 3-different Count                    & 52.00  & 29.00 &  \textbf{88.00} & 39.87 & 41.35 & 44.00 & 47.28 & 25.00  & 23.60 & 34.41 & 45.68 & 35.85 \\

& 1-different \& 2-similar Orientation & 31.00  & 28.00  & 32.99 & 29.32 & 29.81 & 36.00 & \textbf{36.73} & 25.50  & 25.50 & 25.00 & 24.73 & 27.91 \\
& 1-different \& 2-similar Color       & 51.00  & 27.00  & \textbf{67.00} & 36.10 & 43.16 & 59.00 & 58.40 & 20.20  & 25.60 & 55.21 & 31.96 & 31.88 \\
& 1-different \& 2-similar Count       & 24.00  & 27.27  & \textbf{43.43} & 31.34 & 31.67 & 36.00 & 37.04 & 23.80  & 26.40 & 35.71 & 30.67 & 35.85 \\

& Count Progression                    & 50.00  & 13.00  & \textbf{77.00} & 37.20 & 43.48 & 48.00 & 45.10 & 23.40  & 24.20 & 36.96 & 43.04 & 34.82 \\
& Count Arithmetic                     & 52.00  & 22.00  & \textbf{75.76} & 43.20 & 45.00 & 49.00 & 47.03 & 27.00  & 25.20 & 33.33 & 35.96 & 36.63 \\
& Count MinMax                         & 29.59  & 22.00  & \textbf{45.45} & 28.74 & 30.00 & 37.00 & 35.73 & 25.60  & 25.80 & 29.35 & 28.57 & 27.43 \\
&  \mycc \textbf{Average}                     &  \mycc 49.79  &  \mycc 24.12  &  \mycc \textbf{62.29} &  \mycc 37.18 &  \mycc 41.73 &  \mycc 46.75 &  \mycc 47.29 &  \mycc 25.49  &  \mycc 25.65  & \mycc 39.15 &  \mycc 38.19 & \mycc 33.65\\
\midrule
\multirow{6}{*}{\textbf{Level-3}} 
& Independent Color-Object-Orientation & 34.00  & 25.25  & \textbf{38.00} & 27.39 & 26.11 & 32.60 & 33.15 & 21.97 & 26.48 & 19.00 & 21.84 & 27.56 \\
& Independent Count-Object-Color       & 34.00  & 24.00  & \textbf{44.00} & 29.45 & 33.33 & 36.34 & 35.78 & 25.47 & 27.35 & 29.00 & 21.13 & 30.15 \\
& Coupled Color-Orientation     & 24.24  & 24.00  & 16.33 & 26.13 & 25.21 & \textbf{29.43 }& 26.20 & 26.47 &  27.01 & 20.00 & 18.07 & 27.45 \\
& Coupled Color-Count                  & 30.00  & 22.00  & 21.21 & 27.46 & 28.64 & \textbf{33.33} & 27.99 & 26.60 &  29.00 & 28.00 & 27.57 & 21.57 \\
& Spiral Color-Orientation             & 28.00  & 22.00  & \textbf{28.87} & 17.31 & 21.00 & 20.45 & 22.77 & 22.86 & 23.14 & 22.00 & 20.27 & 20.75 \\
& Spiral Color-Count-Object            & \textbf{56.00}  & 30.00  & 54.17 & 28.63 & 30.49 & 36.00 & 37.94 & 25.00 & 27.16 & 33.00 & 27.14 & 27.45\\
&  \mycc \textbf{Average}                     &  \mycc \textbf{34.39}  &  \mycc  23.86  &  \mycc 33.73 &  \mycc 26.31 &  \mycc 27.46 &  \mycc 31.36 &  \mycc 30.64 &  \mycc 24.90 &  \mycc26.88 & \mycc 25.17 &  \mycc 22.67 & \mycc 25.82\\
     \bottomrule[0.1em]
    \end{tabular}}
\caption{Accuracy across VisRes benchmark levels and subtasks under guided prompting with thinking mode enabled (when available).}

\label{sota_low_res}
\end{table*}

VisRes comprises  19,000 evaluation samples distributed across three levels. All tasks use real-world images in a four-choice visual selection format. As shown in \reffig{fig:visres}, Level 1 assesses perceptual grounding through local patch completion and global scene reconstruction under visual degradation. Level 2 (5,956 samples across 12 subtasks) evaluates rule inference over a single visual attribute $3{\times}3$ using Raven-style grids. Level 3 (2,522 samples across 6 subtasks) requires multi-attribute compositional reasoning, with grids governed by concurrent rules spanning multiple visual domains. Together, these levels provide a structured framework for diagnosing visual reasoning capabilities in current models.

\subsection{Task Design and Taxonomy}

\noindent\textbf{Perceptual Completion Tasks:} 
Includes local patch completion and global occlusion. Local tasks match a masked tile to one of four candidates under controlled perturbations (blur, brightness, rotation, edges, orientation), while global tasks require inferring scene structure when 50–80\% of the image is occluded. These tasks assess the robustness of low-level visual representations under degradation.

\noindent\textbf{Single-Attribute Rule Tasks:} These use a $3{\times}3$ Raven-style grid where the missing cell is determined by a row-wise rule applied to one attribute—color, count, or orientation—while others vary freely. Rules include progressions, categorical distributions (3-different or 2-similar–1-different), and simple arithmetic operations, enabling targeted evaluation of single-attribute abstraction.

\noindent\textbf{Multi-Attribute Rules:} These tasks require integrating multiple jointly varying attributes such as color, count, object identity, and orientation. Subtasks span coupled rules, independent multi-attribute rules, and spatially structured patterns like spirals, probing whether models can compose several transformations simultaneously, reflecting the highest level of relational and compositional difficulty in VisRes. 
The detailed taxonomy is provided in the supplementary, expanding the one given in Table~\ref{sota_low_res} and Section~\ref{sec:task_gen}.

\subsection{Image Collection and Preprocessing}
\label{sec:collection}

\noindent\textbf{Level-1:} Our benchmark is built from a pool of natural images collected from multiple sources, including Google Street View scenes for occlusion tasks and public web images for local completion tasks. All images were filtered to exclude harmful or inappropriate content and center-cropped to ensure consistent aspect ratio. For local completion tasks, masked patches of 80×80 px were applied, represented as white tiles with black outlines. The final task is assembled into a single 512×512 composite image that contains both the main image and the four candidate patches (A–D). For global occlusion tasks, each candidate (ground truth and distractors) is stored as an independent 512×512 image, and the query is formed by presenting the occluded version alongside these four options.

\noindent\textbf{Levels- 2 \& 3:} We require per-image labels for count, color, and orientation. We construct these labels using a semi-automated pipeline combining metadata, model-based verification, and targeted manual annotation. For count and color, we begin with an initial label extracted from the textual keyword used during web crawling (e.g., five white dogs). We then verify the count using the Molmo counting model \cite{deitke2025molmo}, retaining only images where the model prediction matches the keyword-derived label. To assess reliability, we manually annotated a random set of 100 images and observed near-perfect agreement between the keyword+Molmo pipeline and human annotations. For color, we apply the same validation strategy by prompting GPT-5 with a constrained attribute-extraction prompt and retaining only images where the model output agrees with the initial metadata. For orientation, metadata is typically unreliable, and models performed poorly (Table~\ref{tab:attributes}). We therefore manually annotated 10k images, assigning each image to one of nine predefined orientation categories. This final annotated data with at least two attributes per image serves as the basis of composition to create raven-like puzzles.
\subsection{Task Generation}
\label{sec:task_gen}

\begin{table*}
\centering
\scriptsize
\setlength{\tabcolsep}{4pt}

\begin{minipage}[t]{0.60\textwidth}
\begin{tabular}{l c c c c c c c c c}
\toprule[0.1em]
Setting & Location & Blur & Brightness & Rotation & Edges  & Global (50\%) & Global (80\%) & \textbf{Avg} \\
\midrule
Original       & 24.3 & 23.9 & 23.7 & 25.5 & 25.1 & 24.9 & 23.9 & \textbf{24.5} \\
Finetuned      & 42.8 & 37.5 & 39.8 & 50.8 & 33.2 & 52.2 & 38.6 & \textbf{43.7} \\
\mycc Human Baseline & \mycc 94.1 & \mycc 84.3 & \mycc 85.6 & \mycc 92.0    & \mycc 82.6 & \mycc 96.1 & \mycc 98.0 & \mycc \textbf{90.4}\\
\midrule
\end{tabular}
\caption{\textbf{Finetuning Qwen2.5-3B on Level~1.} After finetuning, the model performance naturally improves on the Level~1 tasks. However, the performance is well-below human baseline.}

\label{tab:local_completion}
\end{minipage}
\hfill
\begin{minipage}[t]{0.30\textwidth}
\centering
\begin{tabular}{p{2cm} c c}
\toprule
Attribute & \textbf{GPT-4o} & \textbf{GPT-5}  \\
\midrule
Color       & 84.6 & 97.6 \\
Orientation & 39.8 &  49.6 \\
Count       & 72.4 & 94.2 \\
\bottomrule[0.1em]
\end{tabular}
\caption{Frontier-models have high performance on single attributes, but struggle with inferrring orientation.}
\label{tab:attributes}
\end{minipage}

\end{table*}

\textbf{Level-1: Local patch completion.} We construct distractors (i.e., incorrect patches) using two strategies: \textbf{Random Sampling (RS)} and \textbf{DINOv2 Similarity (DS)}. In RS, patches are extracted from non-overlapping regions of the same image without intersecting the masked tile. In DS, 64 non-overlapping candidate patches are uniformly sampled, DINOv2-large embeddings are computed, and the three most similar patches (by cosine similarity) are selected. When image-level augmentations (blur, brightness, rotation, edges) are applied, both the ground-truth and DS distractors are re-extracted at their original coordinates after augmentation to preserve spatial alignment.

We apply the augmentations listed in Level-1 section of Table~\ref{sota_low_res}. We only present results with \textbf{DS} distractors in the main paper as they are perceptually harder. \textbf{RS} results and the actual parameters of the augmentations are detailed in the supplementary.

\noindent\textbf{Level-1: Global occlusion.} These tasks move from the patch-level inference to scene-level inference under heavy information loss. Large portions are masked using uniformly distributed square occluders, covering 50\% or 80\% of the image (cf. ~\reftab{sota_low_res}). Distractors are globally plausible, sharing layout and content but are subtly misaligned, forcing models to infer the exact continuation of the occluded scene. This parallels human amodal completion, where global layout must be inferred despite severe occlusion. The exact distractor sampling strategy and the occluder generation strategy are detailed in the supplementary.

\noindent\textbf{Level-2: Single-Attribute Rule Tasks.} 12 subtasks evaluate single-attribute reasoning across three visual attributes—color, orientation, and count. Each presents a $3\times3$ Raven-style grid~\cite{raven1983manual} with one missing cell, requiring inference of a rule applied to a single target attribute while others vary freely. The missing cell is fixed at position (2,2) to ensure structural consistency and isolate single-attribute effects. The logical structures within rows include:
\begin{itemize}
\item \textbf{Uniform patterns:} all cells within each row share the same attribute value, while the specific value may differ between rows. For example, in \textit{Uniform Color}, each row is uniformly colored (e.g., all red); the same applies for Count and Orientation. 
        
\item \textbf{Distribution Patterns:} the target attribute varies categorically within a row, while all other attributes remain random. Two configurations are used: three distinct values (\textit{3-different}) or a two-same-one-different arrangement (\textit{2-similar 1-different}). For example, in \textit{Color 3-different}, each row contains three distinct colors (e.g., red–green–blue); and the goal is to identify the missing object with the required color. In \textit{Count 2-similar 1-different}, in a single row, two cells share the same number of objects while one differs.

\item \textbf{Progression Patterns:} the number of objects varies monotonically or cyclically within a row (e.g., $1\!\rightarrow\!2\!\rightarrow\!3$ or $3\!\rightarrow\!2\!\rightarrow\!1$). For example, in \textit{Count Progression}, each row follows an increasing or decreasing numerical sequence of object counts.

\item \textbf{Quantitative Operations:} arithmetic relations are applied to object counts within a row, including min–max selection and additive or subtractive operation (e.g., the third image contains the sum, difference, or extremum of object counts from the first two). For example, in \textit{Count Min-Max}, the third cell shows either the smallest or largest number of objects; and in \textit{Count Arithmetic}, it depicts the sum or difference of the preceding two.

\end{itemize}

\noindent\textbf{Level-3: Multi-Attribute Rule Tasks.} 
This level comprises six subtasks extending beyond single-attribute reasoning to test compositional inference across multiple visual dimensions: color, count, orientation, and object identity. Each $3\times3$ grid requires integrating concurrent rules spanning these attributes and combines symbolic, quantitative, and spatial reasoning. The position of the missing cell varies across the grid to prevent positional shortcuts and ensure comparability across grid-wise (i.e., coupled variant), row-wise (i.e. independent variant), and spiral variants. Three categories of logical structure are implemented:

\begin{itemize}
    \item \textbf{Coupled Attribute Rules:} in this setting, relations between attributes are deterministic, where one value uniquely predicts another (e.g., a specific count always paired with a specific color, or left-oriented objects always colored red). \textit{Includes: Coupled: Color-Orientation, Color-Count.}

    \item \textbf{Independent Multi-Rule Compositions:} involves simultaneous reasoning across unlinked dimensions. One variant combines \textit{arithmetic relations on count} (e.g., $1 + 2 = 3$ or $3 - 1 = 2$) with independent color and object patterns, while the other applies three parallel distribution rules for color, object, and orientation within each row. Each attribute can follow either a \textit{3-different} or \textit{2-similar 1-different} categorical pattern independently, creating mixed distributions across dimensions.  \textit{Includes: Independent Count-Object-Color, Independent Color-Object-Orientation.}

    \item \textbf{Spatial-Compositional tasks:} structured transformations are applied along continuous spatial paths where attribute values change jointly or according to independent rules per attribute. In \textit{Spiral Color-Orientation}, orientation follows the spiral traversal direction (left/right/up/down) while colors cycle through a fixed sequence; in \textit{Spiral Color-Object-Count}, color, object identity, and count cycle independently along the spiral path (e.g., colors repeat red–green–blue–yellow, objects alternate mug–ball–car–cat, counts progress $1 \!\rightarrow\! 2 \!\rightarrow\! 3 \!\rightarrow\! 4 \!\rightarrow\! 1 …$). \textit{Includes: Spiral: Color-Orientation.}

\end{itemize}

The options include one correct completion and three procedurally generated distractors that systematically violate the given rule. For color and orientation tasks, distractors alter only the target attribute while preserving spatial structure (e.g., swapping a color from another row). For count-based rules, distractors alter one operand or operation (e.g., reversing progression, swapping addition $\leftrightarrow$ subtraction, or substituting an incorrect extreme). The tasks and their response formats are shown in Fig.~\ref{fig:visres},~\ref{fig:visres_explainer}. Images may be reused across the dataset, but any two tasks share at most two images to promote diversity.

\section{Experiments}

All tasks are presented in a four-choice format, consisting of a main image and four candidate options (A–D). Each sample is accompanied by metadata specifying the subtask type, distractor generation strategy, augmentation parameters, ground-truth coordinates, and the correct answer index. This setup ensures reproducibility and enables systematic analysis of task difficulty across conditions. Performance is measured using accuracy, and results are reported both per subtask and in aggregate to compare model performance at different levels of granularity.

\subsection{Main results}

\noindent \textbf{Models:} We evaluate state-of-the-art vision–language models: Qwen2.5-VL\cite{bai2025qwen2}, Qwen3-VL\cite{yang2025qwen3}, GPT-4o \cite{hurst2024gpt}, GPT-5, InternVL3.5 \cite{wang2025internvl3}, GLM-4.5V\cite{v2507glm}, Kimi-vl \cite{team2025kimi}, and Mimo-VL \cite{xiaomi2025mimo}. These models were selected based on their reported performance on widely used benchmarks.

\noindent \textbf{Prompt design:} Two prompt variants were used across all levels. Generic prompts provided minimal guidance, forcing visual task inference. Guided prompts indicated which visual attribute to attend to (e.g., color, count, orientation) and, when relevant, the relational rule type (e.g., similarity, progression, combination). Both variants maintained identical visual layouts and answer options, differing only in phrasing, allowing systematic comparison between unguided and guided visual reasoning. Full prompt templates are provided in the Supplementary Material. We report results of guided in \reftab{sota_low_res}.

\subsection{Overall performance:} 
\label{sec:performance}
As shown in \reftab{sota_low_res}, we evaluate all models under guided prompting with thinking mode enabled. We set the context length to 32k tokens to allow extended reasoning chains. If a model fails to provide a definitive answer and instead repeats thinking or exceeds the context limit without concluding, we mark the response as incorrect. This accounts for some results falling below random chance (25\%), as certain models, occasionally loop in reasoning without reaching a decision. Performance varies across levels, model classes, and attribute types.

\noindent \textbf{Level-1:} Models exhibit consistent performance patterns across perceptual transformations. Rotation achieves higher accuracy across models (35.42\% GPT-5, 37.5\% Qwen3-4B, 36.05\% Qwen3-30B, 34.38\% Gemini), suggesting that orientation changes may produce more distinctive visual features. Global occlusion at 50\% shows substantial model-dependent variation: Gemini achieves 57.14\%, MiMo-VL-7B 48.35\%, and Qwen3-30B 47.25\%, while GPT-4o achieves 20.88\%. Edge detection and location-based matching yield lower performances, approaching random chance. This suggests that matching based on structural composition or spatial relationships presents greater difficulty than detecting transformation-based changes.

\noindent \textbf{Level-2:} performance varies systematically by attribute type. \textit{Color reasoning} achieves the highest accuracy: Uniform Color reaches 96–97\% for GPT-5 and Gemini, with open-source models achieving 66–91\%. Complex color patterns (3-different, 2-similar-1-different) maintain a good accuracy. \textit{Count reasoning} achieves intermediate performance: top models score 77–90\% on count progression and arithmetic, while open-source models achieve 37–49\%. \textit{Orientation reasoning} yields lower accuracy: Uniform Orientation scores remain 19–30\% across models. Within count-based tasks, arithmetic operations (sum, min, max) yield 10–15 point lower accuracy than simple progressions across models.

\noindent \textbf{Level-3:} multi-attribute integration reduces performance relative to Level-2. \textit{Independent} attribute tasks (where each attribute follows a separate rule) achieve 29–44\% on Count-Object-Color, while \textit{coupled} tasks (where attributes are bound together) achieve 21–33\% on Color-Count. Spiral Color-Count-Object achieves higher performance (56\% GPT-5, 54.17\% Gemini) compared to other Level-3 tasks, while Spiral Color-Orientation achieves 17–28\%. The performance drop on Level-3 indicates increased difficulty when multiple attributes must be jointly reasoned over.

\noindent \textbf{Model-class differences:} Closed-source models (GPT-5, Gemini) achieve higher performance on Level-2 compared to open-source models, while this gap is smaller on Level-1 and Level-3. The reduced gap on Level-1 may indicate that perceptual tasks are less discriminative across model classes, while the reduced gap on Level-3 suggests that compositional reasoning remains challenging regardless of model type. Among open-source models, larger models (Qwen3-30B, Qwen3-32B and Mimo-VL) achieve 38–46\% on Level-2, while smaller models (InternVL-4B/8B, Qwen3-4B/8B) perform closer to the random baseline. This pattern suggests that model capacity correlates with reasoning performance within the open-source model class. Full results across prompting variants (guided vs. generic), few-shot conditions, and thinking mode ablations are provided in the supplementary material.

\subsection{Effect of training}

To establish an upper bound, we recruited five human participants who each completed 200 tasks sampled from across the benchmark, including 20 tasks from each of the eight local completion variants and 20 tasks at 50\% and 80\% occlusion for the global setting. Approximately 60\% of the tasks overlapped between participants to support comparison, with the remainder unique to each individual. Humans achieved an average accuracy of about 91\% across all tasks, performing near ceiling (94–98\%) on location and occlusion variants, with moderate declines under blur (84\%), brightness (85\%), and edge-only conditions (79–85\%). These results confirm that the tasks are perceptually intuitive and solvable for humans while revealing a clear gap relative to current model performance.

A natural question arises of whether these failures simply reflect gaps in the pretraining data. If so, could supervised fine-tuning enable the models to solve these tasks, and reveal which features of are learnable. To investigate this, we adopt a standard supervised SFT setup using Qwen2.5-VL-3B. We restrict our study to Level-1 tasks, where high-quality data can be generated at scale, and create a training set of 100k images per subtask.

\reftab{tab:local_completion} shows that supervised fine-tuning yields consistent gains across all Level-1 subtasks. While the pretrained model exhibit random behaviour (25\% accuracy), fine-tuning improves accuracy by over 19 points on average. The largest gains appear in rotation-sensitive and global-context subtasks, suggesting that geometric cues are the most learnable under direct supervision. Improvements on pixel-based perturbations (blur, brightness, edges) are lower, indicating that robustness to low-level visual variations remains harder to acquire. Despite these gains, overall performance remains far from human levels.

\section{Analysis}

\begin{table}[t]
\centering
\setlength{\tabcolsep}{3pt}
\begin{minipage}[t!]{0.47\columnwidth}
\small   %
\begin{center}
\begin{tabular}{p{1.5cm} cc}
\toprule
Model & Level-2 & Level-3 \\
\midrule
GPT-4.1 & 43.6 & 57.8 \\
GPT-5   & 85.0 & 66.0  \\
\bottomrule
\end{tabular}
\caption{Text-verbalized results for Level-2 and Level-3.}
\label{tab:level23}
\end{center}
\end{minipage}%
\hfill
\begin{minipage}[t!]{0.47\columnwidth}
\small
\begin{center}
\begin{tabular}{l ccc}
\toprule
Setting & 16 & 32 & 48 \\
\midrule
Location & 62.6 & 45.6 & 39.4  \\
Blur     & 59.4 & 46.2 & 37.6\\
Rotation & 47.8 & 42.6  & 40.1 \\
\bottomrule
\end{tabular}
\caption{MAE Level-1 scores for different tile sizes.}
\label{mae_results}
\end{center}
\end{minipage}

\end{table}

Performance on VisRes remains far from optimal across most tasks. We investigate whether failures originate from perceptual grounding, or reasoning operations through controlled experiments: evaluating specialized vision encoders (MAE), varying input resolution, isolating attribute recognition, and testing text-only reasoning.

\subsection{The impact of Thinking-mode}

We study the question: \textit{is explicit thinking required?} We compare standard inference against thinking mode across all three levels. ~\reftab{thinking} shows the effect of enabling versus disabling thinking. Thinking mode consistently improves performance across all models and levels. Notably, open-source models perform near random chance (${\sim}25\%$) without thinking but clearly improve when enabled. The largest improvements occur on Level-2 tasks, where single-attribute reasoning benefits significantly from explicit intermediate steps. GPT-5 demonstrates the strongest overall performance, though the gap between high and low thinking modes remains relatively small, indicating that even minimal reasoning guidance aids visual abstraction.

\begin{table}
    \centering
 \adjustbox{width=0.95\columnwidth}{
\begin{tabular}{l cc cc cc cc} 
 \toprule[0.1em]
   & \multicolumn{2}{c}{\textbf{GPT-5}} 
   & \multicolumn{2}{c}{\textbf{Mimo-VL}}
   & \multicolumn{2}{c}{\textbf{Qwen3-4B}}
   & \multicolumn{2}{c}{\textbf{Qwen3-30B}} \\
 \cmidrule(lr){2-3} \cmidrule(lr){4-5} \cmidrule(lr){6-7} \cmidrule(lr){8-9}
   \textbf{Thinking} & high & low & \cmark & \xmark & \cmark & \xmark & \cmark & \xmark \\
 \midrule[0.1em]
\textbf{Level-1} & 32.61 & 31.43 &  29.22 & 23.91 &  28.17 & 23.16 & 31.20 & 23.60 \\
\textbf{Level-2} & 49.79 & 47.01 &  39.15 & 26.68 &  37.18 & 24.08 & 46.75 & 28.25\\
\textbf{Level-3} & 34.39 & 32.89 &  25.17 & 25.23 &  26.31 & 23.50 & 31.36 & 24.00\\
\bottomrule[0.1em]
\end{tabular}}
\caption{\textbf{Impact of thinking mode on visual reasoning accuracy.} Open-source models (Mimo-VL, Qwen) perform near random without thinking but improve when enabled. \cmark~indicates thinking mode enabled, \xmark~indicates disabled.}
\label{thinking}
\end{table}

\subsection{Why Do Models Fail?}
\begin{table}[t]
\centering
\small
\begin{tabular}{p{3cm}ccc}
 \toprule[0.1em]
\textbf{Resolution} & \textbf{Level~1} & \textbf{Level~2} & \textbf{Level~3} \\
\hline
512$\times$512 & 45.17 & 42.83 & 31.63 \\ 
1024$\times$1024 & 54.01 & \textbf{49.61} & 35.48 \\ 
2048$\times$2048 & \textbf{56.51} & 48.99 & \textbf{40.07} \\ 
\bottomrule[0.1em]
\end{tabular}%
\caption{\small Studying the impact of image resolution on GPT-5. All levels improve with higher number of image tokens.}
\label{tab:accuracy_levels}
\end{table}

The low performances across most subtasks raises a fundamental question: \textit{what is the source of failure?} We identify three potential bottlenecks. First, models may lack sufficient visual resolution to perceive fine-grained details in small patches or grids. Second, even with adequate resolution, they may exhibit perceptual deficits, struggling to count objects accurately, distinguish orientations, or recognize colors consistently. Third, assuming perception is intact, failures may reflect reasoning limitations, an inability to infer and apply abstract rules governing attribute transformations. These failure modes are not mutually exclusive, but they are distinct. By varying resolution, isolating attribute-specific tasks, and analyzing error patterns across rule types, we systematically probe where the perception-to-reasoning pipeline in VLMs breaks. The following subsections present controlled experiments addressing each potential failure mode in turn.

\noindent\textbf{Resolution Impact:} We set the baseline image size to $512^{2}$ pixels, providing sufficient detail to resolve fine-grained visual patterns for an average human. However, VLMs exhibit strong resolution dependence due to their tokenization strategy. To isolate the effect of spatial context, we vary input resolution from $512 \rightarrow 1024  \rightarrow 2048$. ~\reftab{tab:accuracy_levels} reveals that higher resolution consistently improves performance across all complexity levels. At baseline resolution, models achieve 45.17\%, 42.83\%, and 31.63\% on Levels 1, 2, and 3 respectively. Doubling resolution to $1024 \times 1024$ yields substantial gains (54.01\%, 49.61\%, 35.48\%), and further increasing to $2048 \times 2048$ produces the strongest overall results (56.51\%, 48.99\%, 40.07\%). The gains are most pronounced on Level-1 perceptual tasks (\textbf{+11.34} points from $512 \rightarrow 1024$) and Level-3 compositional reasoning (+8.44 points), suggesting that fine-grained visual detail is important for both low-level perception and high-level relational inference. However, even at maximum resolution, performance remains low, indicating that while resolution is a limiting factor, it is not the sole bottleneck—perceptual grounding and reasoning deficits persist even when adequate visual information is available.

\noindent\textbf{Perceptual Grounding:} To isolate perception from reasoning, we construct single-cell attribute recognition tasks. For each grid, we randomly select one cell and ask: \textit{"What is color/count/orientation in this cell?"} with four choices. Models must simply extract a single attribute value. ~\reftab{tab:attributes} shows that GPT-4.1 achieves 84.6\% accuracy on color recognition, 72.4\% on counting, and 39.8\% on orientation. Performance varies substantially across attributes, with a 44.8-point gap between color and orientation. Suggesting limited capability for geometric attribute detection. These results indicate attribute-specific perceptual limitations, with spatial and geometric properties (orientation) presenting greater difficulty than color-based or count.

\noindent\textbf{Pure Reasoning Capacity:} Finally, we test reasoning in isolation by verbalizing tasks entirely as text. Each grid cell is described symbolically (\textit{"3 blue globes"}, \textit{"2 red mugs"}), and answer options are presented as text rather than images. The logical structure is preserved, but vision is eliminated. ~\reftab{tab:level23} shows that GPT-5 achieves 85.0\% on Level-2 (single-attribute rules) but 66.0\% on Level-3 (multi-attribute rules). This 19-point difference indicates that compositional reasoning over multiple simultaneous attributes remains more difficult than single-attribute inference, even in text-only format. 

\noindent\textbf{Findings:} Comparing GPT-5's performance across experiments helps hypothesize few bottlenecks. On Level-2 tasks, text-only accuracy (85.0\%) exceeds visual accuracy (50.0\%); Level-3 shows a similar gap (66.0\% vs. 37.0\%). This modality gap persists across task complexity, indicating that limitations arise from visual feature extraction rather than reasoning capacity. Perceptual grounding results (~\reftab{tab:attributes}) show differential performance: color (84.6\%) and count (72.4\%) recognition are above 70\%, while orientation (39.8\%) remains near chance (25\%). Resolution increases from $512 \times 512$ to $2048 \times 2048$ improve Level-2 accuracy from 42.8\% to 49.0\% (~\reftab{tab:accuracy_levels}), but still leave a gap relative to text-only performance.

These results indicate three failure points: (1) partial perceptual deficits, particularly for spatial attributes; (2) incomplete visual feature extraction at available resolutions; (3) limited integration of visual features into reasoning operations. Text-only performance demonstrates that reasoning mechanisms exist, but are not effectively applied when operating on visual input. The limitation is in visual-to-symbolic translation rather than logical inference.

\subsection{Can vision encoders solve Level-1?}
We attempt to evaluate a vision encoder natively trained on similar task, hence, masked autoencoder (MAE)~\cite{He2021MaskedAA} is the choice. We design a retrieval task that tests whether reconstructed patches can correctly identify their corresponding ground truth regions. Given an input image $\mathbf{I} \in \mathbb{R}^{H \times W \times 3}$, we randomly sample a square tile region $\mathbf{C}_0 \subset \mathbf{I}$ of size $s \times s$ pixels and mask it during encoding. The MAE model reconstructs this masked region, producing $\hat{\mathbf{C}} \in \mathbb{R}^{s \times s \times 3}$. We generate $k-1$ distractor crops $\{\mathbf{C}_1, \ldots, \mathbf{C}_{k-1}\}$ from the same image with varying spatial overlap relative to $\mathbf{C}_0$, where $k=4$ in our experiments.

\noindent\textbf{Similarity Metrics.}
We evaluate reconstruction quality using two complementary metrics. For pixel-level similarity, we compute the $L_2$ distance in RGB space:
\begin{equation}
d_{\text{pixel}}(\hat{\mathbf{C}}, \mathbf{C}_i) = \|\hat{\mathbf{C}} - \mathbf{C}_i\|_2 = \sqrt{\sum_{h,w,c} (\hat{\mathbf{C}}_{hwc} - \mathbf{C}_{i,hwc})^2}
\end{equation}

\noindent\textbf{Retrieval Accuracy.} For each reconstruction $\hat{\mathbf{C}}$, we rank the $k$ crops $\{\mathbf{C}_0, \mathbf{C}_1, \ldots, \mathbf{C}_{k-1}\}$ by their similarity to $\hat{\mathbf{C}}$. We report accuracy across different tile sizes in ~\reftab{mae_results}. The results demonstrate that pixel-level reconstruction quality degrades as tile size increases, with accuracy dropping from 62.6\% for $16^{2}$  to 39.4\% for $48^{2}$ tile size. Notably, even at the largest tile size, performance remains significantly above random chance (25\% for $k=4$), indicating that MAE reconstructions preserve meaningful spatial information. The inverse relationship between tile size and retrieval accuracy suggests that reconstruction fidelity decreases for larger masked regions, likely due to the increased complexity and ambiguity in predicting larger image areas.

\section{Conclusion}

We introduced VisRes, a benchmark designed to diagnose visual reasoning capabilities in naturalistic imagery across perceptual, relational, and compositional task settings. Our evaluation of state-of-the-art vision–language models reveals systematic failures across all task types. Despite strong performance on traditional vision-language benchmarks, current VLMs perform near random baseline on most VisRes tasks, exposing fundamental limitations in visual reasoning when linguistic guidance is removed. VisRes allows us to isolate where these failures occur. Level-1 perceptual tasks show that models struggle with basic visual completion, with foundational models scoring around 50\%, a task an average human solves with ease. Fine-tuning yields moderate improvements but performance remains far below human baselines, indicating that perceptual deficits extend beyond data exposure. As the task demands increase from single-attribute reasoning (Level-2) to multi-attribute composition (Level-3), performance drops, suggesting that perceptual limitations hinder the model’s ability to infer and combine attribute-level rules. Controlled experiments identify the primary bottleneck: a persistent gap between text-only and vision-based reasoning. Models demonstrate strong reasoning capabilities when operating on symbolic text descriptions, yet fail to apply these same capabilities when reasoning must be extracted from visual input. This modality gap persists across resolutions and task types, indicating that limitations arise from visual-to-symbolic verbalization rather than logical inference itself. Attribute-specific perceptual tests confirm differential deficits, with geometric and spatial features proving particularly difficult to extract reliably.

{
\small
\bibliography{refferences}
\bibliographystyle{refferences_style}
}

\newpage
\newpage

\appendix

\setcounter{table}{0}
\setcounter{figure}{0}
\renewcommand{\thetable}{A.\arabic{table}}
\renewcommand{\thefigure}{A.\arabic{figure}}

\section{Task generation: Level 1}

\textbf{Level-1: Local patch completion.}

In this setting, a tile of the main image is masked, and the model must select the correct patch from four candidates. Tasks are generated under two strategies: \textbf{Random Sampling (RS)} and \textbf {DINOv2 Similarity (DS)}. RS distractors are sampled from unrelated regions within the same image. In DS tasks, 64 patches are uniformly sampled, embedded with DINOv2-large, and the top-3 most similar to the ground-truth patch are selected. When augmentation is applied, distractors are re-sampled at matching coordinates to preserve consistency. This enforces content-based similarity while allowing perceptual uncertainty. The only exception is the edge-image setting, where similarity search is applied directly in edge space after converting the image to an edge map.

We implement three categories of local patch tasks:

\begin{itemize}[leftmargin=*]
    \item \textbf{Natural-image tasks:}
    \begin{itemize}
        \item \textbf{Location (RS):} Non-augmented; distractors sampled randomly from other regions.
        \item \textbf{Location (DS):} Non-augmented; distractors chosen via similarity on the original image.
        \item \textbf{Blur (DS):} Main image blurred with Gaussian kernels, $\sigma = 3$--$20$ px.
        \item \textbf{Brightness (DS):} Main image gamma-corrected in RGB space, with $\gamma = 0.10$--$0.25$ for darkening and $\gamma = 3.5$--$5.0$ for brightening.
        \item \textbf{Rotation (DS):} Main image rotated $45$--$90^\circ$ clockwise or counterclockwise, and distractors are re-sampled at the corresponding coordinates.
    \end{itemize}
    
    \item \textbf{Edge-image tasks:}
    \begin{itemize}
        \item \textbf{Edges (RS):} Main image and candidates converted to Canny edge maps (50--150 thresholds), filtered for $\geq 2\%$ edge pixels, $\geq 75\%$ coverage, and $\geq 10$ connected components, then dilated with a $2\times 2$ kernel. Distractors sampled randomly.
        \item \textbf{Edges (DS):} Main image converted to edge map before similarity search; the top-3 most similar edge patches are chosen as distractors, ensuring confusion arises from contour similarity alone.
    \end{itemize}
    
    \item \textbf{Same-patch rotation:}
    \begin{itemize}
        \item \textbf{Rotation same patch (SP):} Main image is rotated $45$--$90^\circ$, and all candidates come from the same masked location. Distractors are produced by applying extra rotations to the ground-truth patch, making orientation the only discriminative cue while keeping content and context identical.
    \end{itemize}
\end{itemize}

\noindent\textbf{Level-1: Global Occlusion.}
Global occlusion tasks extend completion from patch-level to scene-level inference under heavy information loss. Large portions of the image are masked using uniformly distributed square occluders, covering 50\% to 80\% of the image. Square tile sizes are drawn from $\{8, 16, 24, 32 \text{ px}\}$, with overlap producing variable fragmentation. Candidates consist of the ground-truth unoccluded image and three distractors drawn from neighboring frames in Google Street View--style sequences, offset by $10$--$20^\circ$ along the same street trajectory. Distractors are globally plausible, sharing layout and content but subtly misaligned, forcing models to infer the exact continuation of the occluded scene. 

\begin{figure*}[t]
    \centering
    \includegraphics[width=\textwidth]{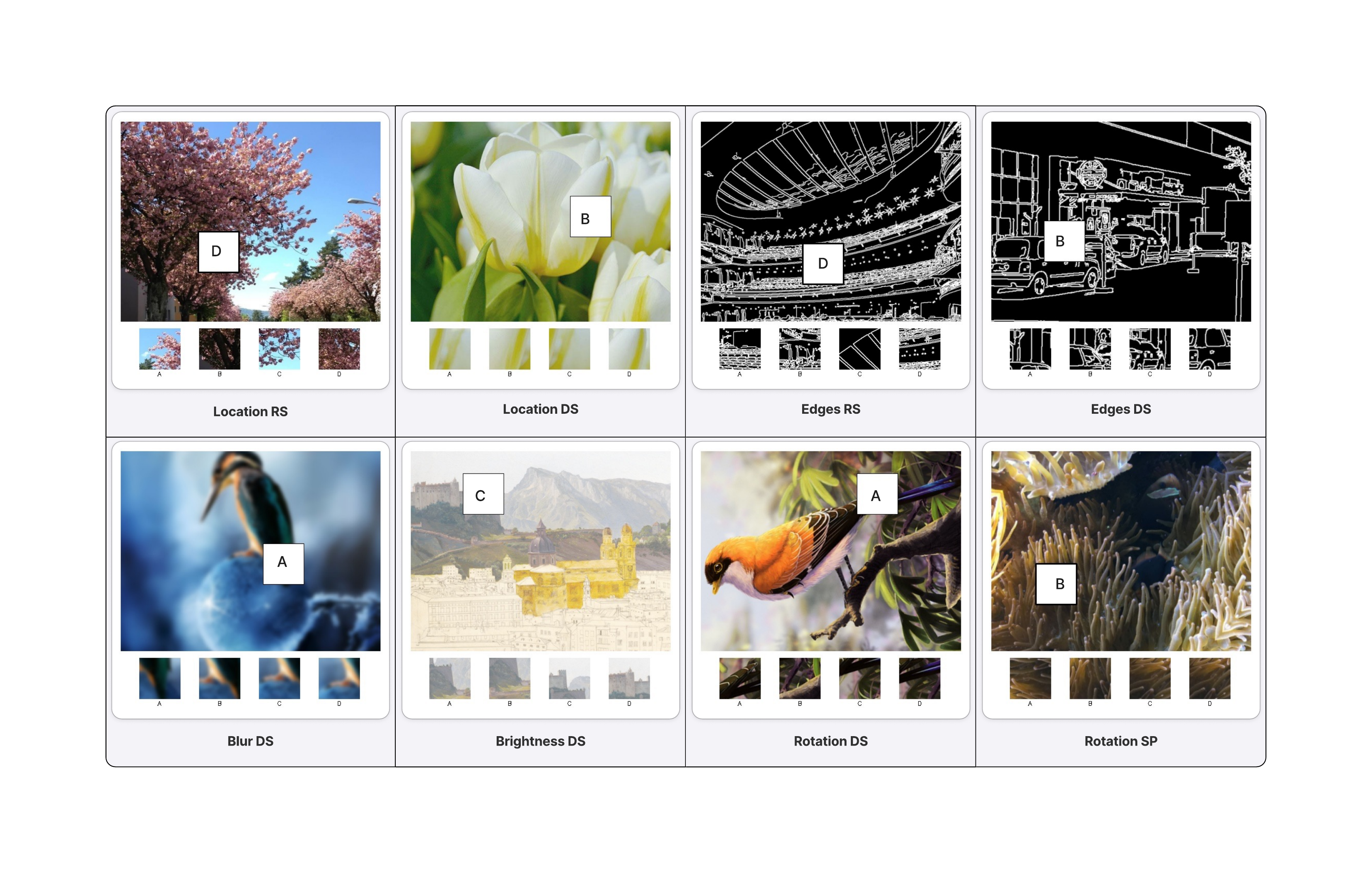}
    \caption{\textbf{Examples of Level-1 perceptual completion tasks.} Each task masks a local region of the main image, and the model must infer the missing content by comparing visual evidence around the blanked area with four candidate patches. The eight task variants isolate different perceptual cues: Location (RS), Location (DS), Edges (RS), Edges (DS), Blur (DS), Brightness (DS), Rotation (DS), Rotation (SP).}
    
    \label{fig:l1-local}
\end{figure*}

\begin{figure*}[t]
    \centering
    \includegraphics[width=\textwidth]{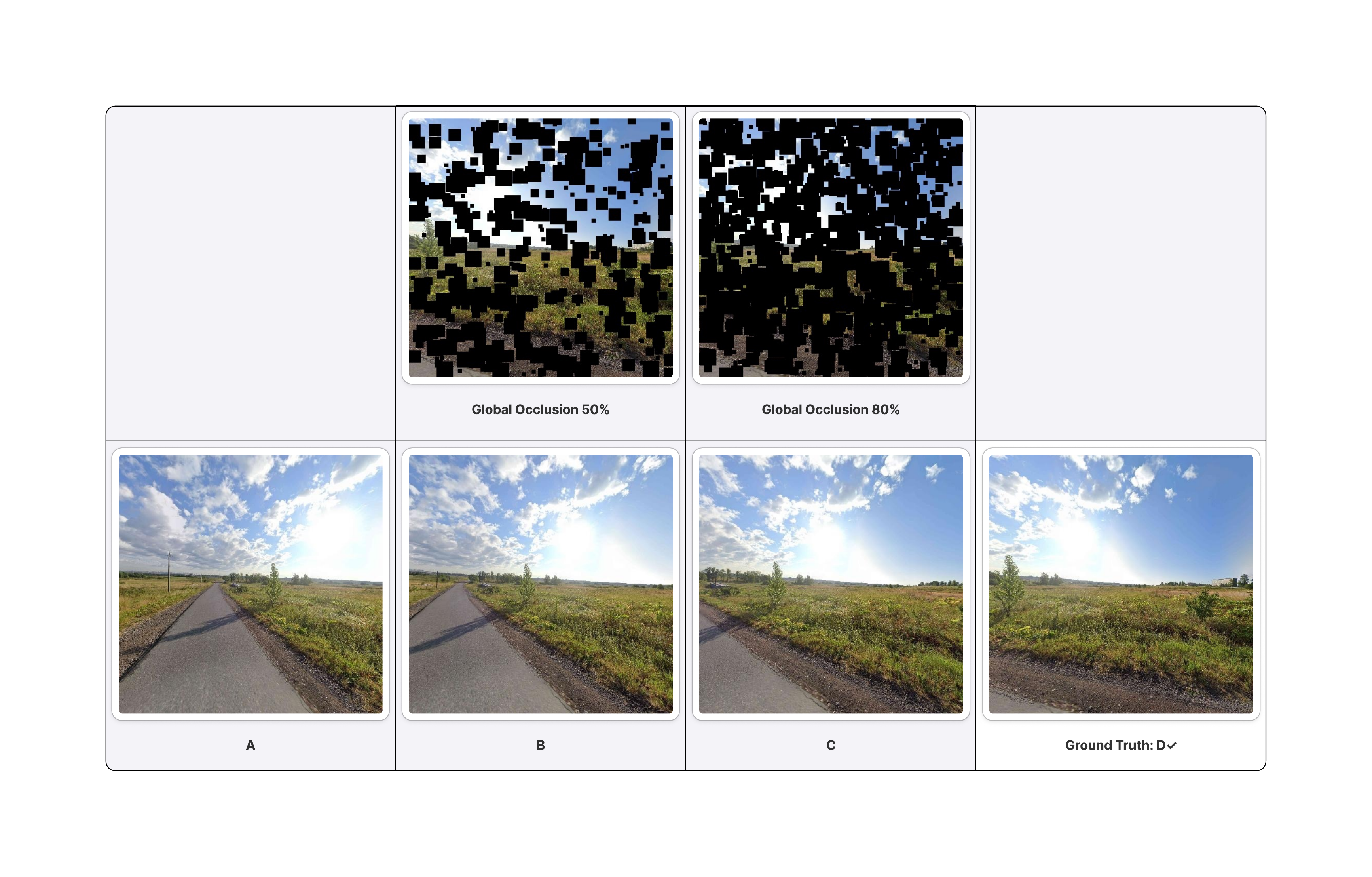}
    \caption{\textbf{Global occlusion tasks in Level-1}, where 50–80\% of the scene is masked. Unlike local completion, the model must reason about the broader structure and context of the scene to recover the missing content. Only one candidate image (A–D) corresponds to the original, unoccluded scene.}
    
    \label{fig:l1-global}
\end{figure*}

\section{Prompt Design Across Levels}

Each task in the benchmark is evaluated with two prompt formulations:
\emph{generic} prompts, which provide no guidance about relevant visual features, and \emph{guided} prompts, which explicitly reference them (i.e., color, orientation, number of objects, object identity).
The visual input, answer format (multiple choice A–D), and layout of the images remain identical in both cases; only the wording changes.
This setup enables a controlled comparison between tasks that rely solely on implicit visual inference and those that explicitly highlight salient visual attributes, assessing whether models benefit from targeted visual guidance in their reasoning process.

\vspace{0.5em}
\subsection*{Level-1 -- Perceptual Completion}

Level~1 uses natural images with partial occlusion. In local completion, a single square region is masked; global completion removes a substantial region of the scene. In both cases, the model must identify which of four candidate patches restores the missing content.

The following generic prompt is used without reference to any specific visual cue (e.g., local patch completion):

\begin{quote}\itshape
You are presented with a main image in which a square region is blanked out (occluded).\\
Below the image are four candidate patches labeled A, B, C, and D, arranged horizontally from left to right.\\
Exactly one patch correctly restores the missing region of the main image.\\
Task:\\
1. Examine the area surrounding the blank region in the main image to understand its context, structure, and visual layout.\\
2. Compare this information with each of the four candidate patches.\\
3. Choose the patch that best completes the missing area, restoring the continuity and coherence of the scene.\\
4. Reply only with the single letter of your chosen answer: A, B, C, or D.
\end{quote}

\textbf{Guided prompts} share the same introductory text above and differ only in the \texttt{Task} instructions. 
Below is an example of a real guided variant (e.g., Location task):

\begin{quote}\itshape
Task:\\
1. Focus on local visual cues around the blank area — such as color, texture, object edges, and lighting direction.\\
2. Compare these clues with each candidate patch to determine which one best aligns structurally and contextually with the main image.\\
3. Select the patch that fits seamlessly into the blank region, maintaining alignment and structural consistency with the surrounding image.\\
4. Reply only with the single letter of your chosen answer: A, B, C, or D.
\end{quote}

Other guided variants differ in the emphasis of Step~1: some instruct robustness to blur and ask for coarse spatial matching; others ask the model to ignore lighting differences and focus on geometric cues. Additional versions highlight contour continuity or rotation. The task format and interface remain constant across all variants.

\vspace{0.8em}
\subsection*{Level-2 -- Single-Attribute Grid Reasoning}

Level~2 uses a \(3\times3\) grid with one missing image. The solution follows a single visual attribute (e.g., color, count, or orientation) that varies systematically across each row. A generic prompt introduces the puzzle as follows:

\begin{quote}\itshape
You are presented with a visual reasoning puzzle consisting of a 3×3 grid of images with one image missing in the bottom-right corner of the grid.\\
Below the grid are four candidate options, labeled A, B, C, and D, arranged horizontally from left to right.\\
Exactly one option correctly completes the grid.\\
Task:\\
1. The logical relationship between images is defined within each row, and the same rule or transformation pattern applies consistently across all rows in the grid.\\
2. Examine how the images change within each row.\\
3. Identify the option that follows the same rule and correctly completes the grid.\\
4. Reply only with the single letter of your chosen answer: A, B, C, or D.
\end{quote}

Guided versions use exactly the same introductory text as above and differ only in the \texttt{Task} section. 
Below is an example for the \texttt{Color} task types:

\begin{quote}\itshape
Task:\\
1. The logical relationship between images is defined within each row, and the same rule or transformation pattern applies consistently across all rows in the grid.\\
2. Study the grid and observe how colors change, repeat, or contrast within each row.\\
3. Identify the option that best fits this color-based rule and correctly completes the grid.\\
4. Reply only with the single letter of your chosen answer: A, B, C, or D.
\end{quote}

Other Level~2 guided tasks follow the same format, altering the visual feature emphasized in Step~2. 
Count-based prompts refer to numerical variation within each row, 
while orientation-based versions highlight directional changes or rotations of objects. 
All task interfaces---including image arrangement and answer format---remain constant across generic and guided versions, 
differing only in the level of visual feature specification.

\vspace{0.8em}
\subsection*{Level-3 -- Multi-Attribute Reasoning}

Level~3 evaluates compositional reasoning over multiple visual attributes simultaneously, 
such as color, object identity, count, and orientation. 
Unlike Level~2, the missing cell may appear in any position within the grid, 
and the governing rule may operate row-wise, grid-wise, or be defined along a spatial trajectory such as a spiral pattern. 
Models must therefore integrate multiple attributes jointly rather than deducing a single rule or transformation. An example of a generic prompt introduces the grid-wise puzzle as follows:

\begin{quote}\itshape
You are presented with a visual reasoning puzzle consisting of a 3×3 grid of images, where one image is missing in one of the nine positions.\\
Below the grid are four candidate options labeled A, B, C, and D, arranged horizontally from left to right.\\
Exactly one option correctly completes the grid.\\
Task:\\
1. The grid follows a uniform logical rule, which governs the relationships between the images across the entire grid.\\
2. The rule may involve multiple visual attributes that change consistently across the grid, following the same underlying logic.\\
3. Carefully analyze how the attributes vary in each image and across the grid to infer the common rule that governs them.\\
4. Choose the option that best follows this rule and completes the grid in a logically consistent way.\\
5. Reply only with the single letter of your chosen answer: A, B, C, or D.
\end{quote}

\textbf{Guided} versions reuse the same introductory text above and differ in the \texttt{Task} instructions.  
Below is a guided example from the dataset (e.g., Independent Count-Object-Color task):

\begin{quote}\itshape
Task:\\
1. The grid follows a row-wise logical rule, where multiple visual attributes change systematically across each row.\\
2. The same rule or transformation pattern applies consistently across all rows in the grid.\\
3. In this puzzle, focus on the following attributes: color, number of elements (count), and object type, which may vary independently or in relation to one another.\\
4. Carefully analyze how these attributes vary across each row, identify the common rule of change, and determine how it applies to the row with the missing image.\\
5. Choose the option that best satisfies the row rule and completes the grid in a logically consistent way.\\
6. Reply only with the single letter of your chosen answer: A, B, C, or D.
\end{quote}

Other Level~3 guided prompts follow the same format while modifying only the attribute(s) referenced in Step~3. 
Depending on the task type, prompts emphasize multiple attributes at once (e.g., color–count, color–orientation), requiring joint compositional reasoning. 
In spatial variants, the relevant attributes must be tracked along a positional trajectory (such as a spiral), while distribution-based or arithmetic tasks require following systematic patterns across rows or across the entire grid. 
Thus, guided prompts differ only in the visual feature emphasized, whereas the input layout and answer interface remain identical across all Level~3 variants.

\begin{figure*}[t]
    \centering
    \includegraphics[width=\textwidth]{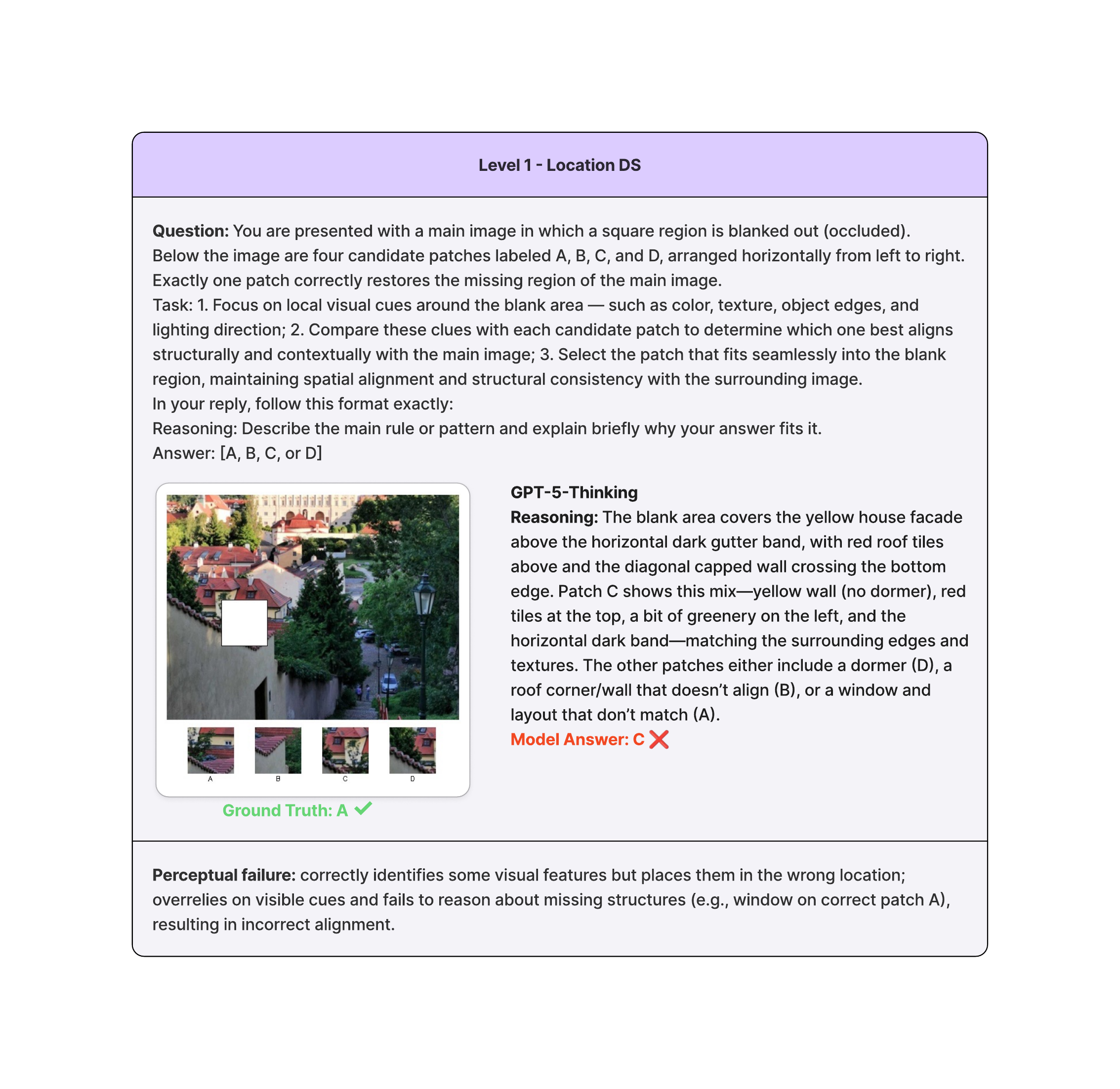}
    \caption{\textbf{Level-1 Location (DS): Prompt and Model Response Example}}
    \label{fig:l1-location_ds}
\end{figure*}

\begin{figure*}[t]
    \centering
    \includegraphics[width=\textwidth]{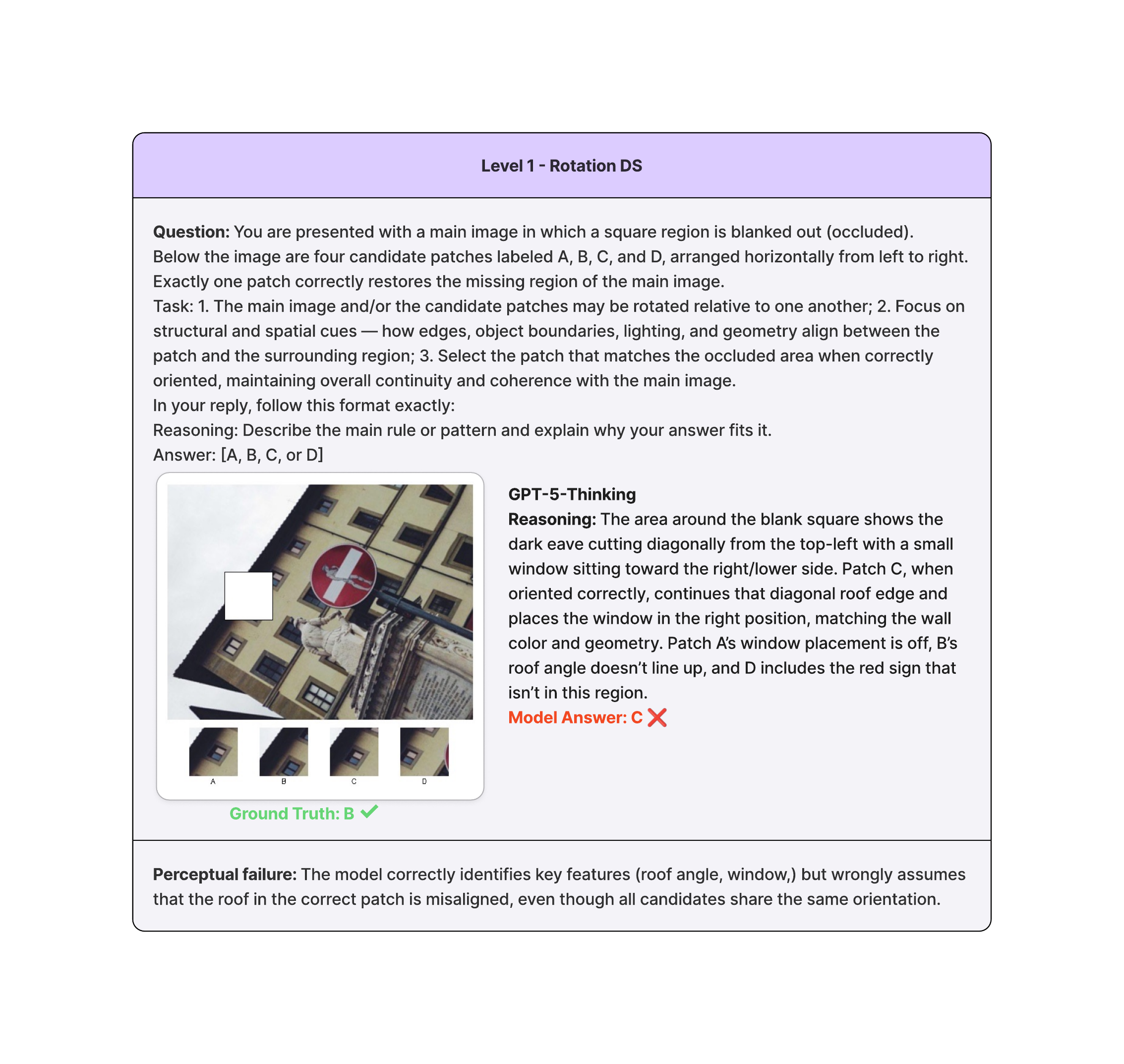}
    \caption{\textbf{Level-1 Rotation (DS): Prompt and Model Response Example}}
    \label{fig:l1-rotation_ds}
\end{figure*}

\begin{figure*}[t]
    \centering
    \includegraphics[width=\textwidth]{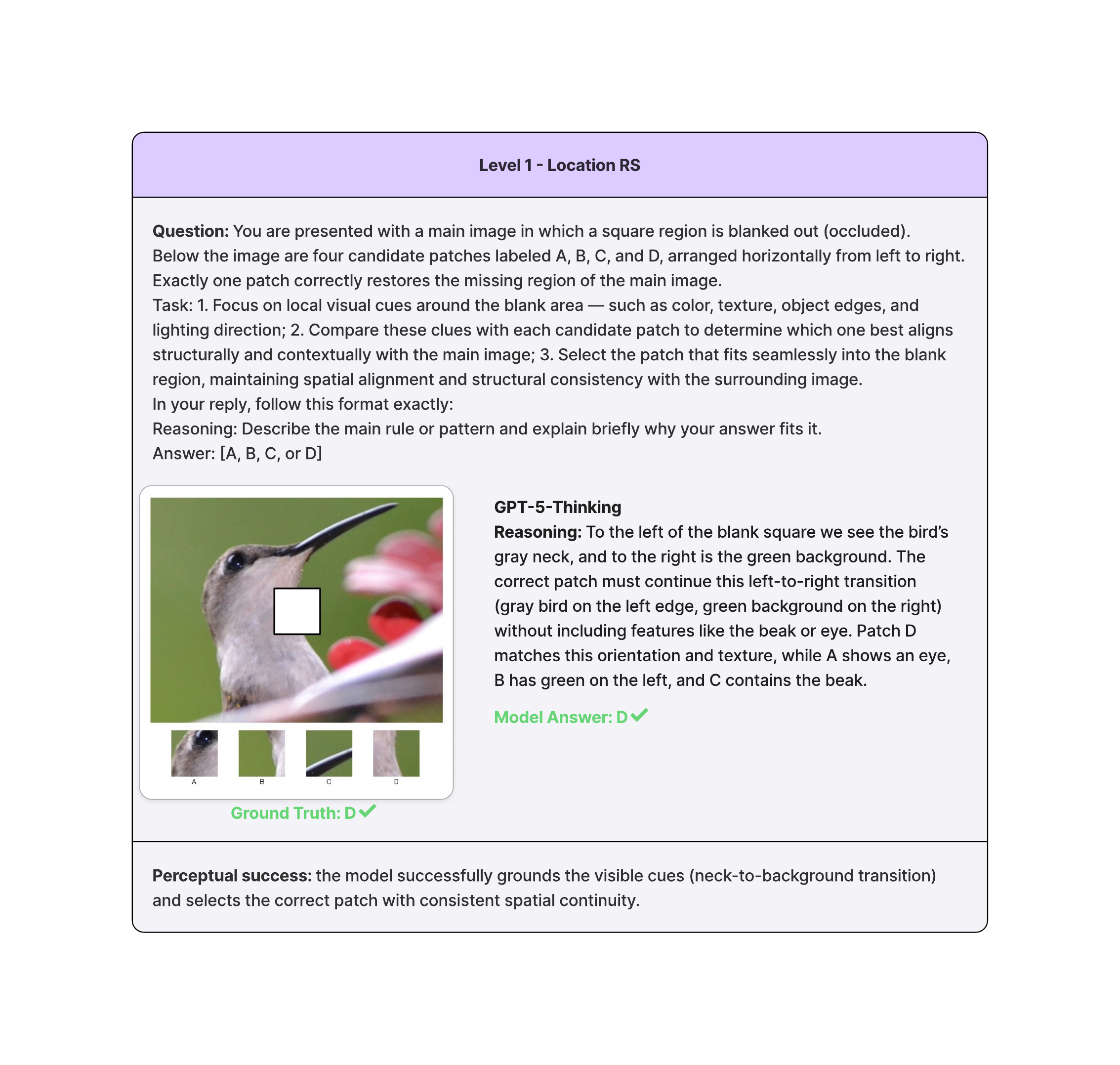}
    \caption{\textbf{Level-1 Location (RS): Prompt and Model Response Example}}
    \label{fig:l1-location_rs}
\end{figure*}

\begin{figure*}[t]
    \centering
    \includegraphics[width=\textwidth]{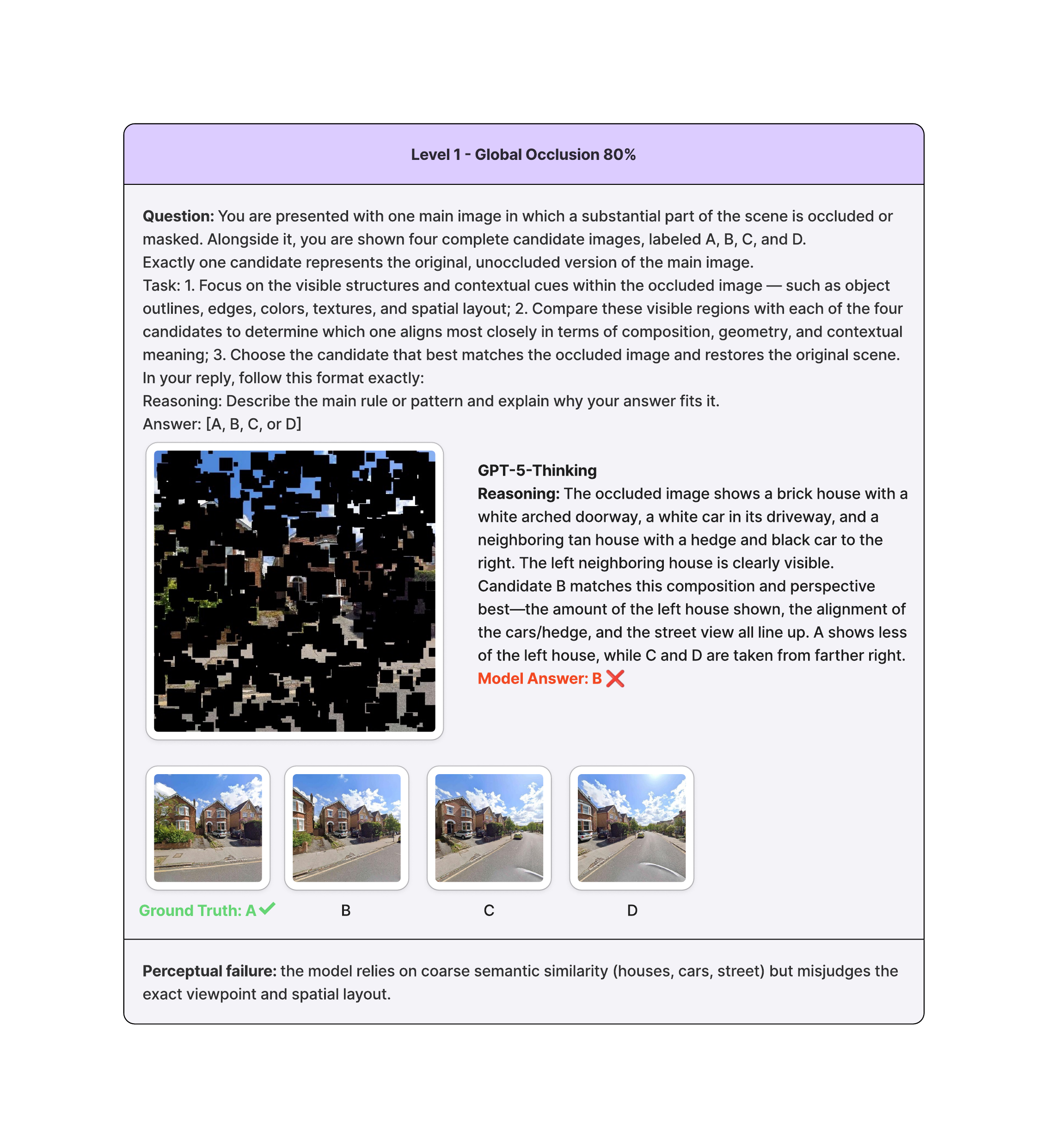}
    \caption{\textbf{Level-1 Global Occlusion 80\%: Prompt and Model Response Example}}
    \label{fig:l1-global_80}
\end{figure*}

\begin{figure*}[t]
    \centering
    \includegraphics[width=\textwidth]{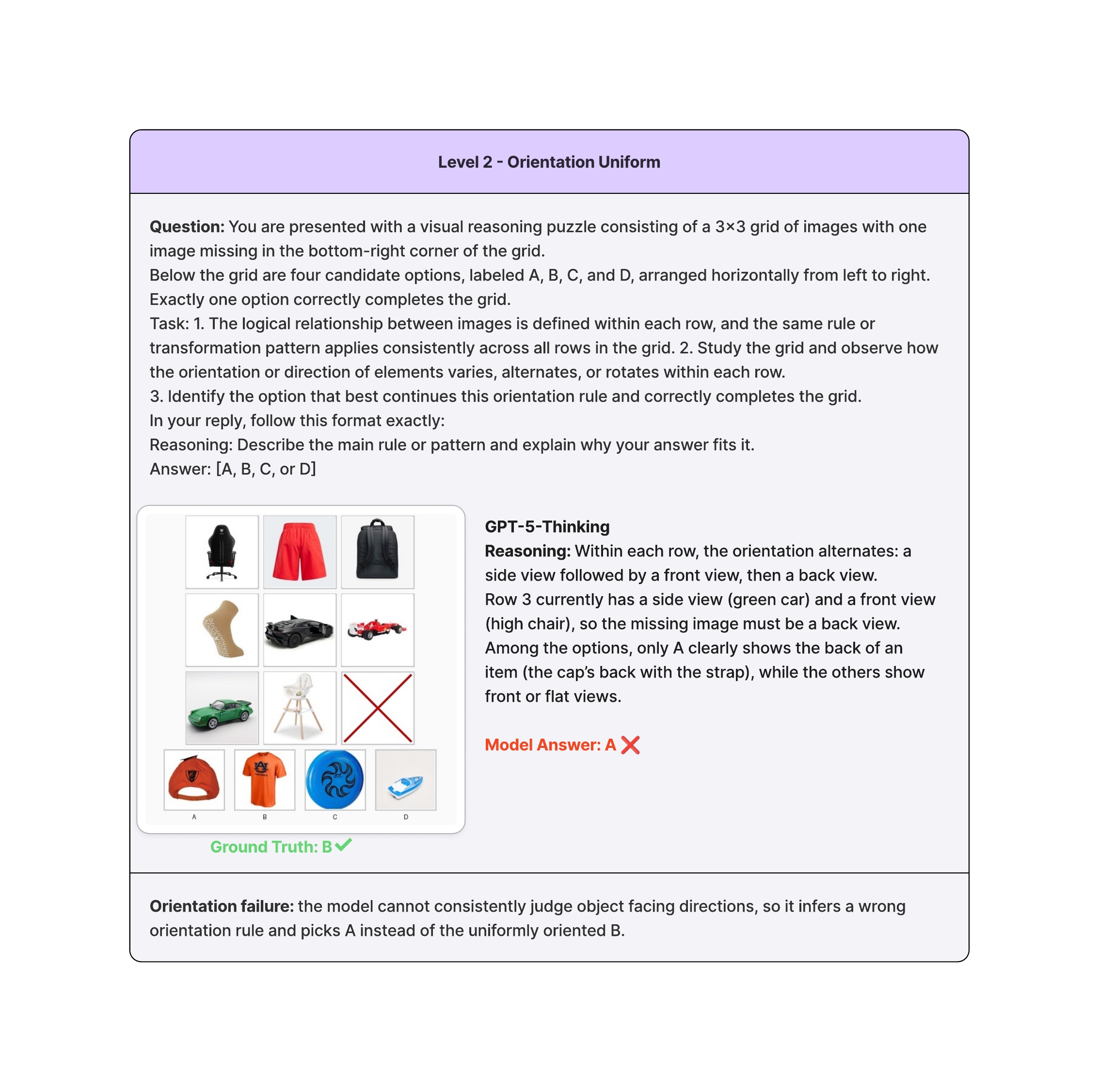}
    \caption{\textbf{Level-2 Orientation Uniform: Prompt and Model Response Example}}
    \label{fig:l2-orient_uniform}
\end{figure*}

\begin{figure*}[t]
    \centering
    \includegraphics[width=\textwidth]{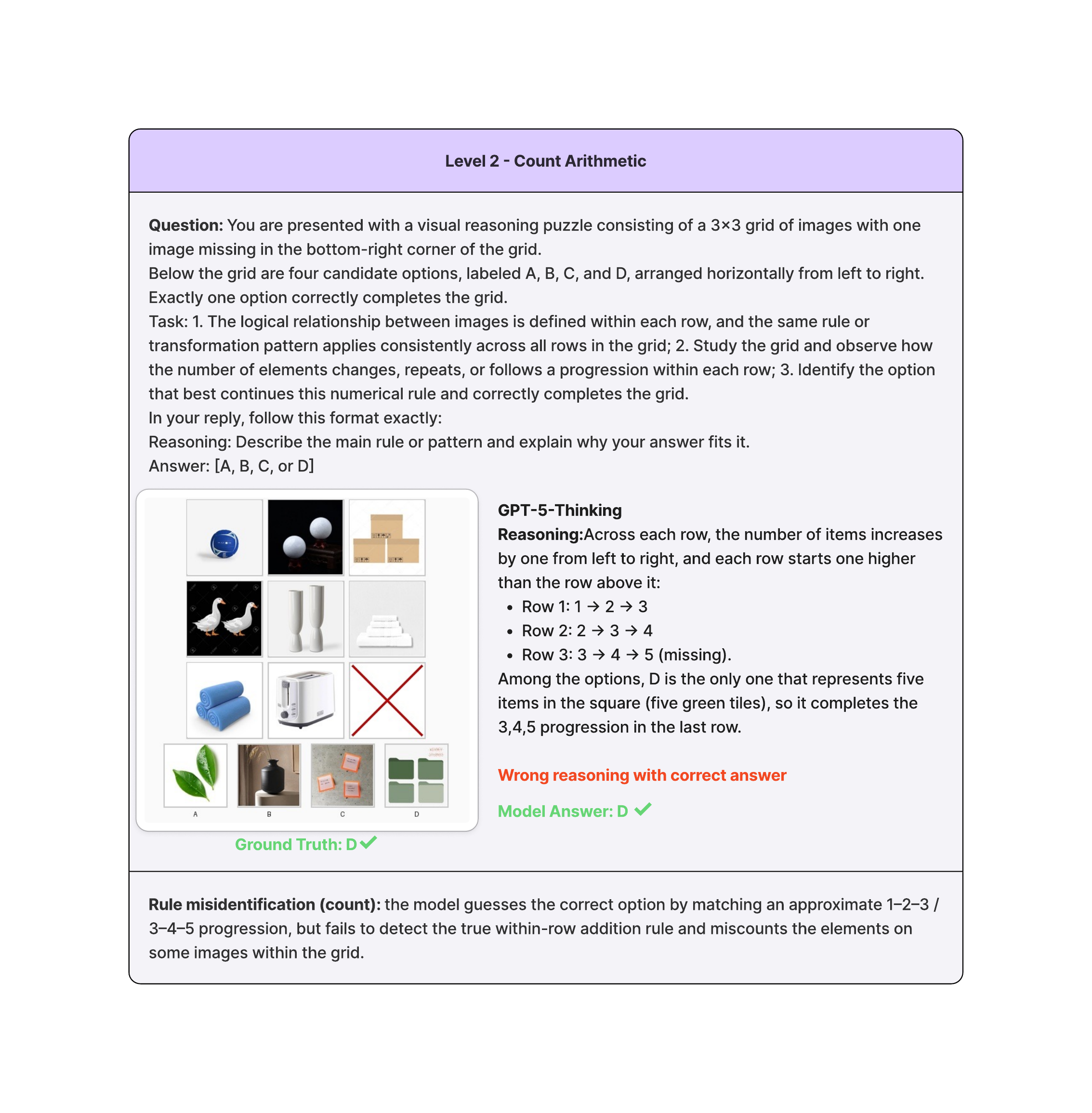}
    \caption{\textbf{Level-2 Count Arithmetic: Prompt and Model Response Example}}
    \label{fig:l2-count_arit}
\end{figure*}

\begin{figure*}[t]
    \centering
    \includegraphics[width=\textwidth]{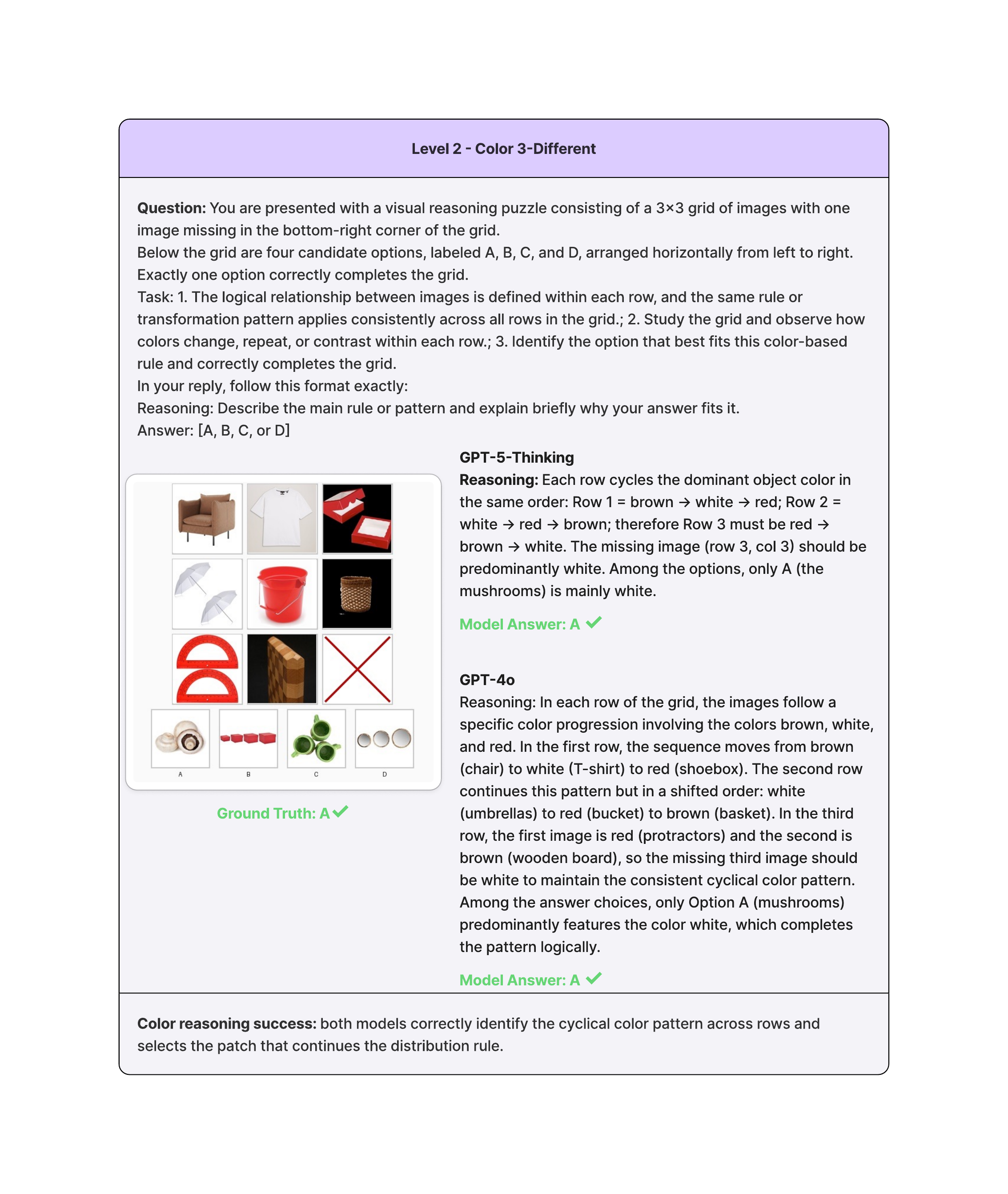}
    \caption{\textbf{Level-2 Color Distribution (3-different): Prompt and Model Response Example}}
    \label{fig:l2-color_3diff}
\end{figure*}

\begin{figure*}[t]
    \centering
    \includegraphics[width=\textwidth]{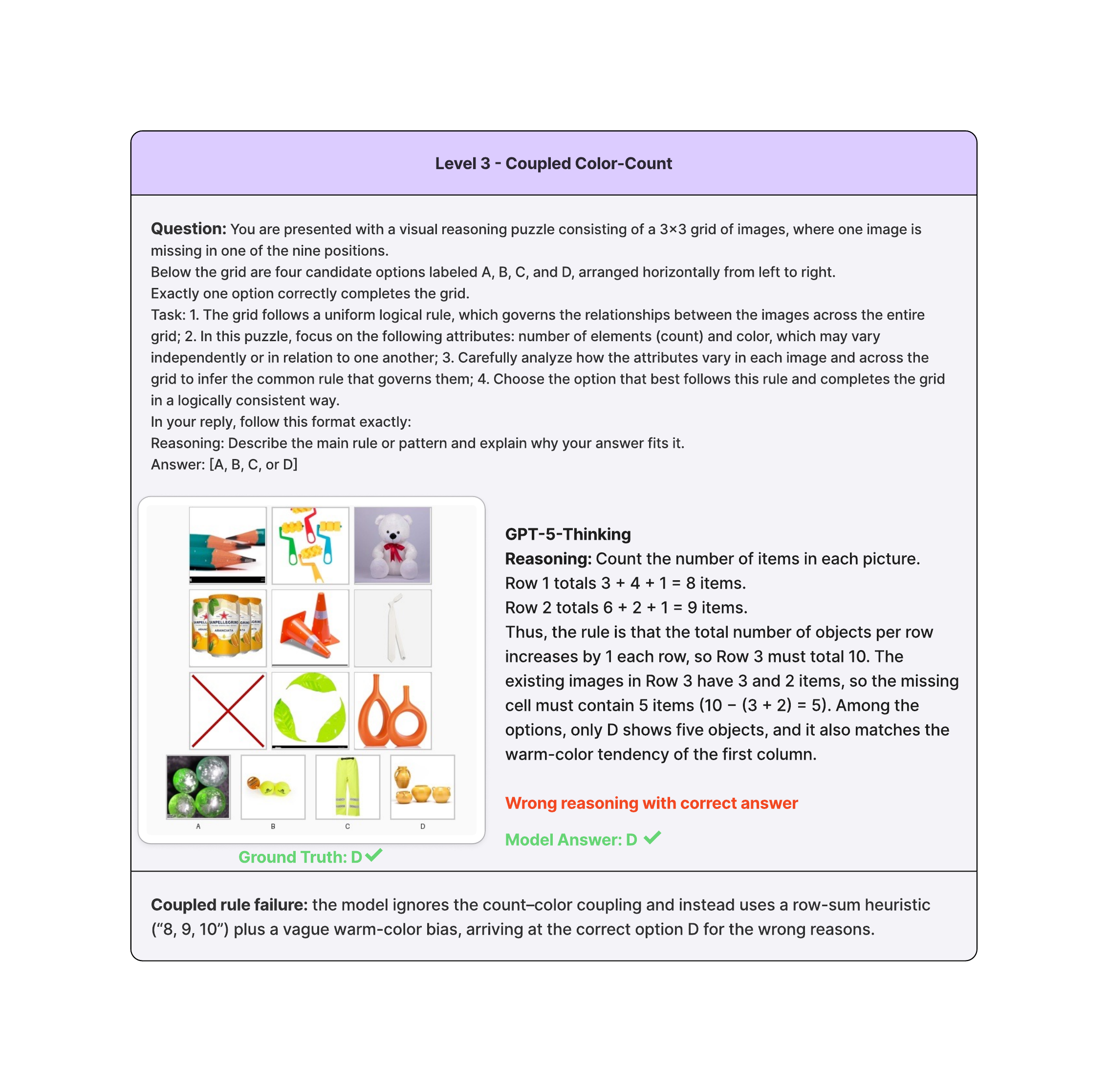}
    \caption{\textbf{Level-3 Coupled Color-Count: Prompt and Model Response Example}}
    \label{fig:l3-coupled}
\end{figure*}

\begin{figure*}[t]
    \centering
    \includegraphics[width=\textwidth]{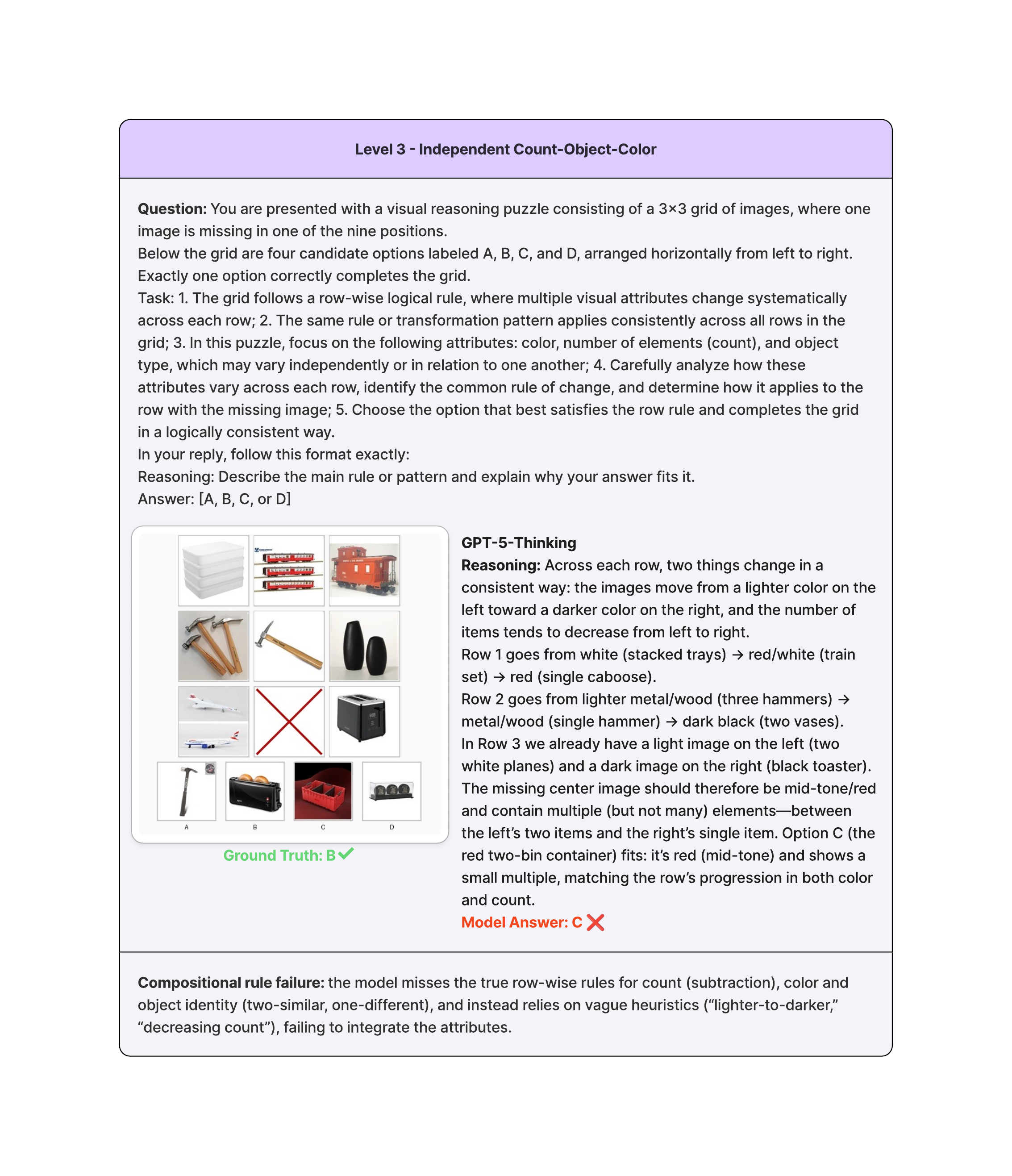}
    \caption{\textbf{Level-3 Independent Count-Object-Color: Prompt and Model Response Example}}
    \label{fig:l3-ind}
\end{figure*}

\begin{figure*}[t]
    \centering
    \includegraphics[width=\textwidth]{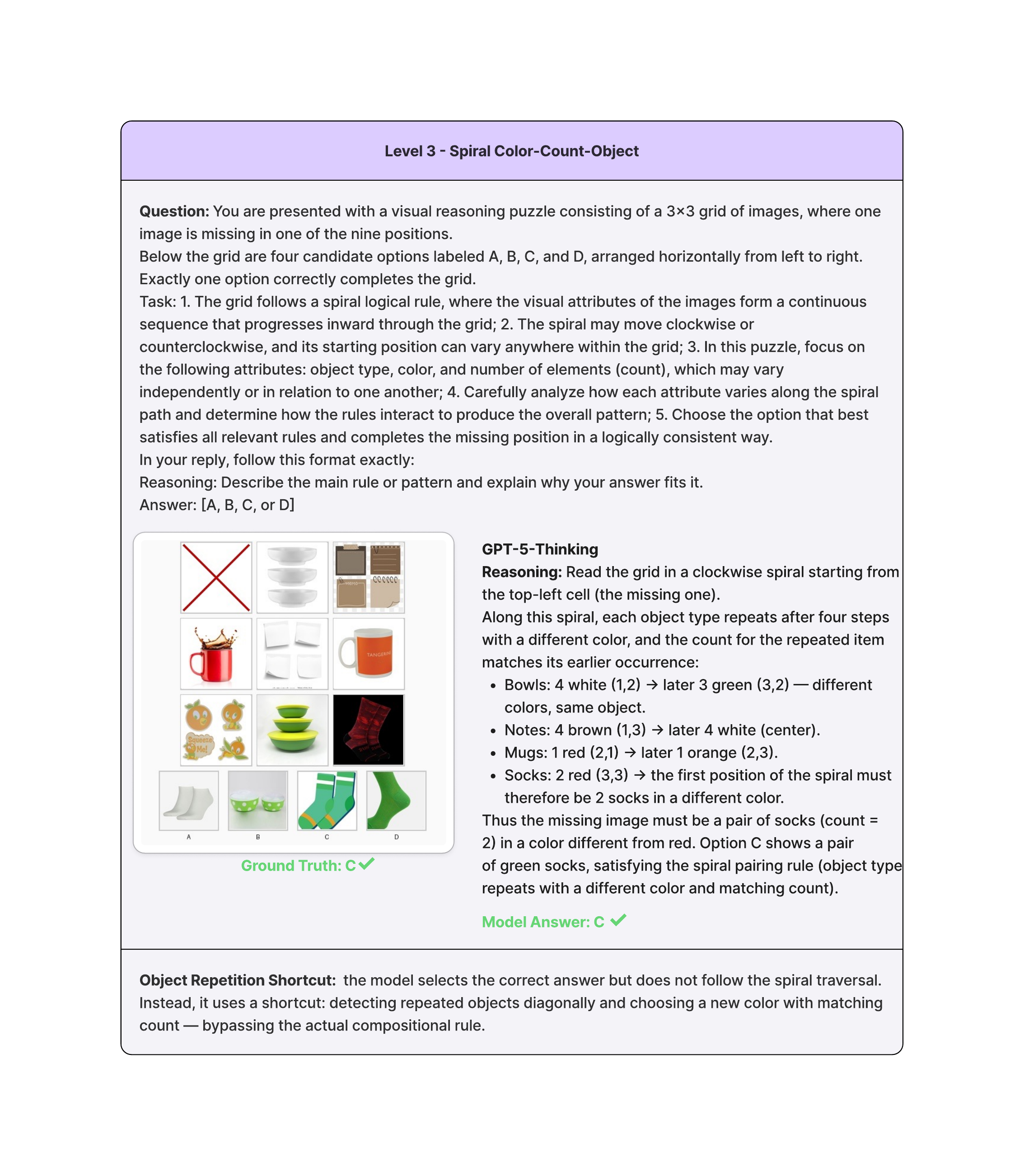}
    \caption{\textbf{Level-3 Spiral Color-Count-Object: Prompt and Model Response Example}}
    \label{fig:l3-spiral_count}
\end{figure*}

\begin{figure*}[t]
    \centering
    \includegraphics[width=\textwidth]{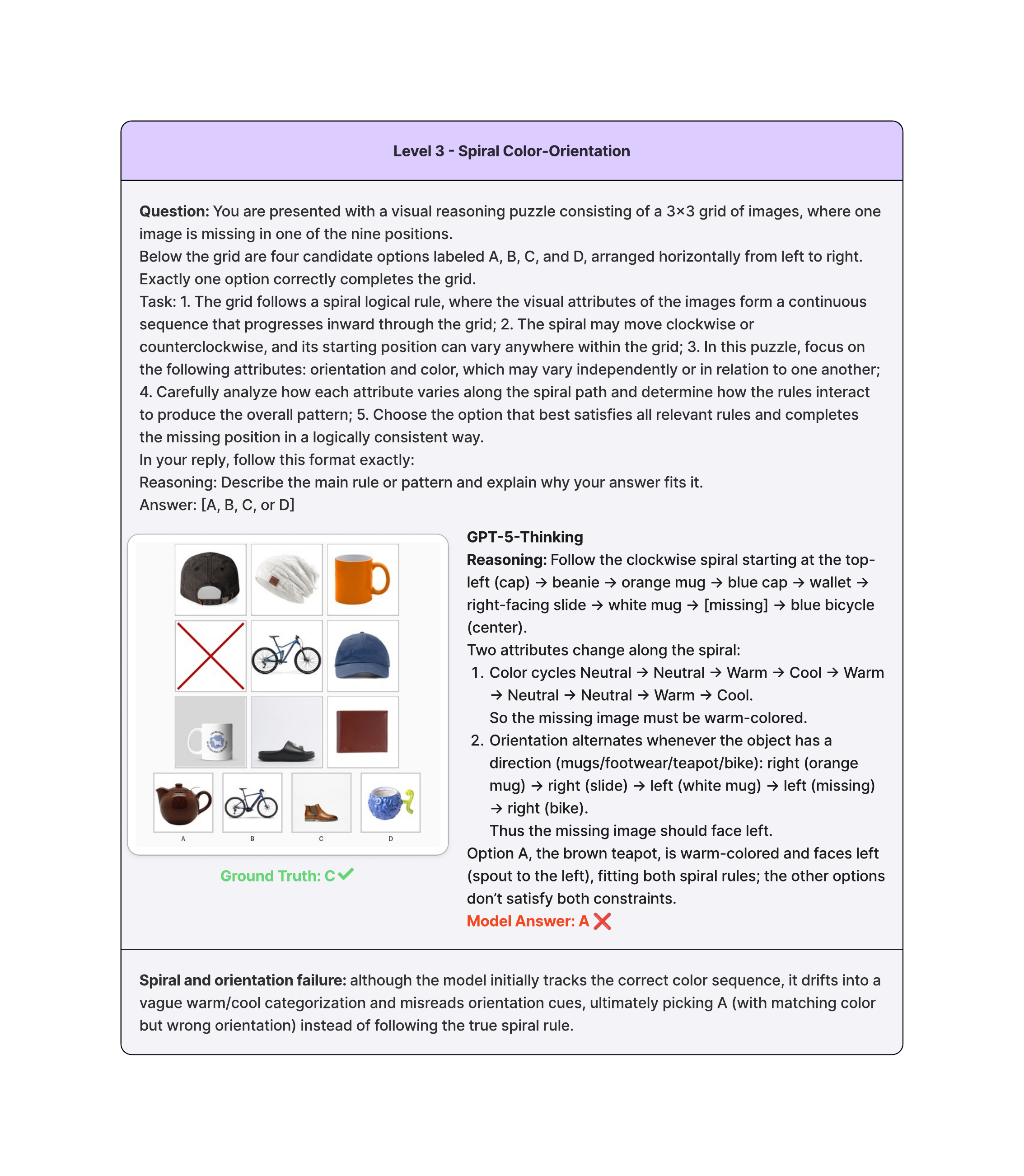}
    \caption{\textbf{Level-3 Spiral Color-Orientation: Prompt and Model Response Example}}
    \label{fig:l3-spiral_orient}
\end{figure*}

\section{Performance details}

This section presents detailed performance analyses—including DS/RS difficulty, guided versus generic prompting, reasoning effort, and few-shot evaluations—that collectively support and extend the empirical findings reported in the main paper.

\begin{table*}[!htbp] 
    \centering
    \setlength{\tabcolsep}{10pt}
 \adjustbox{width=0.7\textwidth}{
\begin{tabular}
{l l c c c c c c c } 
 \toprule[0.1em]

   & \textbf{Setting} 
 & \rotatebox{70}{\textbf{GPT-5}}  
 & \rotatebox{70}{\textbf{GPT-4o} }  
 & \rotatebox{70}{\textbf{Gemini-2.5} } 
 & \rotatebox{70}{\textbf{Qwen3-VL-4B }}  
 & \rotatebox{70}{\textbf{Qwen3-VL-8B }}  
 & \rotatebox{70}{\textbf{Qwen3-VL-30B }}  
 & \rotatebox{70}{\textbf{{Qwen3-VL-32B}}}\\
\toprule
\multirow{6}{*}{\textbf{DS Tasks}} 
& Edges      & \textbf{27.17} & 23.91 & 25.00 & 16.67 & 24.18 & 25.00 & 20.22 \\
& Location   & 23.71 & 20.62 & 26.00 & 23.16 & 22.73 & 22.40 & \textbf{27.27} \\
& Rotation   & 35.42 & 26.04 & 34.38 & \textbf{37.50} & 34.04 & 36.05 & 33.33 \\
& \textbf{Average} 
& \textbf{28.77} & 23.52 & 28.46 & 25.78 & 26.98 & 27.82 & 26.94 \\
\midrule

\multirow{6}{*}{\textbf{RS Tasks}}
& Edges      & 44.14 & 30.70 & \textbf{45.10} & 27.86 & 38.30 & 40.92 & 41.70 \\
& Location   & 60.70 & 37.74 & \textbf{67.80} & 41.50 & 43.50 & 48.65 & 55.10 \\
& Rotation   & \textbf{29.63} & 26.43 & 28.10 & 28.26 & 27.60 & 29.50 & 23.20 \\
& \textbf{Average}
& 44.82 & 31.62 & \textbf{47.00} & 32.54 & 36.47 & 39.69 & 40.00 \\
\bottomrule
    \end{tabular}}
\caption{Accuracy across VisRes benchmark DS vs RS tasks  with thinking mode enabled (when available).}

\label{ds_and_rs}
\end{table*}

\paragraph{Random Sampling (RS) and DINOv2 Similarity (DS):} As shown in \reftab{ds_and_rs}, the contrast between Random Sampling (RS) and DINOv2 Similarity (DS) tasks highlights a substantial gap in perceptual difficulty: models achieve much higher accuracy under RS, where distractors come from unrelated regions and can be rejected through coarse contextual cues, whereas DS markedly lowers performance by presenting DINOv2-selected distractors that are perceptually similar to the target patch. This consistent drop across all architectures indicates that current VLMs rely heavily on shallow texture or boundary heuristics and struggle when required to distinguish between patches with close semantic or structural alignment.

\begin{table*}[!htbp]
\centering
\setlength{\tabcolsep}{6pt}
\adjustbox{width=0.97\textwidth}{
\begin{tabular}{l| 
c c | c c |
c c |c c |
c c |c c}
\toprule[0.1em]
\textbf{Setting}
& \multicolumn{2}{c|}{\rotatebox{70}{\textbf{GPT-5}}}
& \multicolumn{2}{c|}{\rotatebox{70}{\textbf{GPT-4o}}}
& \multicolumn{2}{c|}{\rotatebox{70}{\textbf{Gemini}}}
& \multicolumn{2}{c|}{\rotatebox{70}{\textbf{Qwen3-VL-30B}}}
& \multicolumn{2}{c|}{\rotatebox{70}{\textbf{Kimi}}}
& \multicolumn{2}{c}{\rotatebox{70}{\textbf{MiMo}}} \\
\cmidrule(lr){2-3}
\cmidrule(lr){4-5}
\cmidrule(lr){6-7}
\cmidrule(lr){8-9}
\cmidrule(lr){10-11}
\cmidrule(lr){12-13}
&
\textbf{GU}
& \textbf{GE}
& \textbf{GU}
& \textbf{GE}
& \textbf{GU}
& \textbf{GE}
& \textbf{GU}
& \textbf{GE}
& \textbf{GU}
& \textbf{GE}
& \textbf{GU}
& \textbf{GE} \\
\toprule[0.1em]

Edges                                & \cellcolor[HTML]{EA9999}27.17 & 26.31            & 23.91            & 21.87            & 25.00                         & 25.80                         & 25.00                         & 23.70                         & 23.08           & 24.00            & 22.30           & 21.90            \\
Location                             & 23.71                         & 22.80            & 20.62            & 18.80            & \cellcolor[HTML]{EA9999}26.00 & 25.10                         & 22.40                         & 21.40                         & 18.09           & 23.40            & 25.77           & 21.40            \\
Rotation                             & 35.42                         & 33.41            & 26.04            & 27.69            & 34.48                         & 33.00                         & 36.05                         & \cellcolor[HTML]{EA9999}38.00 & 24.20           & 26.30            & 29.17           & 20.80            \\
Brightness                           & 25.26                         & 22.68            & 27.27            & 23.25            & 27.37                         & 25.10                         & \cellcolor[HTML]{EA9999}29.67 & 21.90                         & 28.57           & 25.30            & 27.37           & 22.20            \\
Blur                                 & \cellcolor[HTML]{EA9999}31.18 & 28.66            & 25.26            & 24.62            & 26.32                         & 28.00                         & 25.11                         & 25.90                         & 30.20           & 23.80            & 26.32           & 22.80            \\
Global\_Occ\_50                      & 42.86                         & 41.43            & 20.88            & 17.84            & \cellcolor[HTML]{EA9999}57.14 & 54.00                         & 40.45                         & 35.10                         & 32.61           & 25.50            & 48.35           & 38.11            \\
Global\_Occ\_80                      & 31.10                         & 29.80            & 32.61            & 17.60            & \cellcolor[HTML]{EA9999}36.96 & 33.40                         & 27.27                         & 25.10                         & 27.78           & 25.10            & 30.43           & 25.95            \\
\midrule
\hspace{6mm}\textbf{Level 1 Average}                            & \textbf{31.11}                         & \textbf{29.77}            & \textbf{23.86}            & \textbf{21.07 }           & \cellcolor[HTML]{EA9999}\textbf{33.28} & \textbf{32.50}                         & \textbf{31.20}                         & \textbf{27.05 }                         & \textbf{26.26}           & \textbf{25.26}            & \textbf{29.22}           &\textbf{ 25.27}            \\
\midrule
Orientation Uniform                  & 22.22                         & 21.58            & 25.25            & 24.22            & 26.53                         & \cellcolor[HTML]{EA9999}27.13 & 25.00                         & 24.00                         & 20.20           & 18.93            & 19.19           & 21.18            \\
Color Uniform                        & 96.00                         & 93.60            & 21.00            & 23.89            & \cellcolor[HTML]{EA9999}97.00 & 89.60                         & 88.00                         & 70.00                         & 45.95           & 42.40            & 78.95           & 66.20            \\
Count Uniform                        & 61.00                         & 54.23            & 25.00            & 24.25            & \cellcolor[HTML]{EA9999}90.91 & 88.40                         & 59.00                         & 57.00                         & 51.48           & 47.56            & 52.75           & 45.45            \\
3-different Orientation              & \cellcolor[HTML]{EA9999}34.69 & 24.75            & 24.00            & 23.72            & 31.96                         & 28.40                         & 17.00                         & 24.00                         & 19.00           & 25.80            & 24.24           & 25.60            \\
3-different Color                    & \cellcolor[HTML]{EA9999}88.00 & 72.14            & 26.00            & 23.6             & 69.00                         & 38.80                         & 65.00                         & 61.90                         & 39.47           & 35.80            & 46.46           & 37.45            \\
3-different Count                    & 52.00                         & 56.39            & 29.00            & 28.2             & \cellcolor[HTML]{EA9999}88.00 & 64.60                         & 44.00                         & 39.60                         & 35.85           & 36.20            & 34.41           & 34.35            \\
1-different \& 2-similar Orientation & 31.00                         & 28.28            & 28.00            & 24.12            & 32.99                         & 28.92                         & \cellcolor[HTML]{EA9999}36.00 & 35.51                         & 27.91           & 23.09            & 25.00           & 23.80            \\
1-different \& 2-similar Color       & 51.00                         & 26.31            & 27.00            & 25.00            & \cellcolor[HTML]{EA9999}67.00 & 27.80                         & 59.00                         & 51.32                         & 31.88           & 30.60            & 55.21           & 36.65            \\
1-different \& 2-similar Count       & 24.00                         & 25.25            & 27.27            & 25.36            & \cellcolor[HTML]{EA9999}43.43 & 41.00                         & 36.00                         & 32.19                         & 35.85           & 33.20            & 35.71           & 33.19            \\
Count Progression                    & 50.00                         & 47.78            & 13.00            & 23.12            & \cellcolor[HTML]{EA9999}77.00 & 69.60                         & 48.00                         & 44.00                         & 34.82           & 35.00            & 36.96           & 32.00            \\
Count Arithmetic                     & 52.00                         & 40.08            & 22.00            & 21.49            & 75.76                         & \cellcolor[HTML]{EA9999}77.00 & 49.00                         & 42.71                         & 36.63           & 34.20            & 33.33           & 32.85            \\
Count Min-Max                        & 29.59                         & 29.23            & 22.00            & 20.86            & 45.45                         & \cellcolor[HTML]{EA9999}47.80 & 37.00                         & 32.53                         & 27.43           & 25.23            & 29.35           & 24.79            \\
\midrule
\hspace{6mm}\textbf{Level 2 Average}                              & \textbf{49.79}                         & \textbf{43.30}            & \textbf{24.12}            & \textbf{23.99}            & \cellcolor[HTML]{EA9999}\textbf{62.29 }& \textbf{52.42}                         & \textbf{46.75   }                      & \textbf{42.90 }                        & \textbf{33.65}           & \textbf{32.33}            & \textbf{39.15}           & \textbf{34.46}            \\
\midrule
Independent Color-Object-Orientation & 34.00                         & 31.71            & 25.25            & 23.94            & \cellcolor[HTML]{EA9999}38.00 & 32.96                         & 32.60                         & 30.42                         & 27.56           & 25.14            & 19.00           & 28.73            \\
Independent Count-Object-Color       & 34.00                         & 32.14            & 24.00            & 20.27            & \cellcolor[HTML]{EA9999}44.00 & 42.17                         & 36.34                         & 33.78                         & 30.15           & 26.04            & 29.00           & 26.04            \\
Coupled Color-Orientation            & 24.24                         & 22.44            & 24.00            & 24.20            & 16.33                         & 24.06                         & \cellcolor[HTML]{EA9999}29.43 & 25.00                         & 27.45           & 26.81            & 20.00           & 26.81            \\
Coupled Color-Count                  & 30.00                         & 21.88            & 22.00            & 22.25            & 21.21                         & 21.80                         & \cellcolor[HTML]{EA9999}33.33 & 16.64                         & 21.57           & 23.69            & 28.00           & 23.69            \\
Spiral Color-Orientation             & 28.00                         & 31.09            & 22.00            & 24.86            & 28.87                         & \cellcolor[HTML]{EA9999}36.29 & 20.45                         & 22.24                         & 20.75           & 24.78            & 22.00           & 24.78            \\
Spiral Color-Count-Object            & \cellcolor[HTML]{EA9999}56.00 & 44.59            & 30.00            & 26.35            & 54.00                         & 54.74                         & 36.00                         & 29.00                         & 27.45           & 26.09            & 33.00           & 26.09            \\
\midrule
\hspace{6mm}\textbf{Level 3 Average}                               & \textbf{34.39 }                        & \textbf{30.64}            & \textbf{23.86}            & \textbf{23.65 }           & \textbf{33.73}                         & \cellcolor[HTML]{EA9999}\textbf{35.34} & \textbf{31.36}                         & \textbf{26.18}                         & \textbf{25.82}           & \textbf{25.43}            & \textbf{25.17 }          & \textbf{26.02}            \\
\midrule
\textbf{Average (All Levels)}                 & \textbf{40.35}                         & \textbf{36.22}            & \textbf{24.47 }           & \textbf{23.01}            & \cellcolor[HTML]{EA9999}\textbf{46.84} & \textbf{42.35 }                        & \textbf{37.78 }                        & \textbf{34.16}                         &\textbf{ 29.73}           & \textbf{28.56}            & \textbf{33.21}           & \textbf{29.68 }           \\

\bottomrule[0.1em]
\end{tabular}}
\caption{Accuracy across VisRes benchmark levels and subtasks under guided and generic prompting. \textbf{GU}: Guided, \textbf{GE}: Generic}
\label{guided_and_generic}
\end{table*}

\paragraph{Impact of Prompt Type on Performance}: Guided prompts consistently improve performance across almost all models and task categories by explicitly directing attention to the relevant visual attributes—such as color, count, or orientation—while generic prompts offer no such cueing (\reftab{guided_and_generic}). This benefit is especially pronounced in Level 2 attribute-based reasoning tasks, where models often gain 10–40 points under guided prompting, indicating that VLMs rely heavily on textual scaffolding when inferring structured visual rules. By contrast, improvements under guided prompting are modest for low-level perceptual tasks such as edges, blur, or global occlusion, suggesting that guidance cannot compensate for weak visual grounding. Overall, the guided vs. generic comparison demonstrates that while models can follow explicit visual reasoning instructions effectively, their performance remains limited when the underlying perceptual signal is ambiguous or fine-grained.

\paragraph{Effect of Few-Shots}: \reftab{few_shots} shows that few-shot examples substantially improve performance on reasoning-driven tasks, while offering only modest gains on low-level perceptual tasks. Across Levels 2 and 3, models benefit strongly from the added demonstrations, often gaining large margins as the examples clarify both the puzzle structure and the underlying attribute rules. In contrast, Level 1 perceptual tasks exhibit smaller or inconsistent improvements, indicating that additional context cannot compensate for weaknesses in fine-grained visual discrimination. Overall, the few-shot results suggest that current VLMs leverage examples primarily to strengthen symbolic rule inference rather than to enhance perceptual grounding, reinforcing the divide between high-level reasoning and low-level visual understanding.

\paragraph{Effect of Enhanced Reasoning}: Models show clear but uneven benefits from increased reasoning effort, with deeper reasoning improving performance mainly on structured visual tasks (in Table \reftab{reasoning}, Level 2 and Level 3 accuracies rise across most models) while producing only small changes on Level 1 perceptual tasks. Because these low-level tasks depend on fine-grained visual grounding rather than symbolic reasoning, adding step-by-step explanations does little to overcome perceptual ambiguity. The table also reveals several cases where models arrive at the correct answer despite flawed or hallucinatory reasoning, suggesting that higher reasoning effort enhances coherence of the explanation more than the underlying perceptual ability.
\begin{table*}[!htbp]
\centering
\setlength{\tabcolsep}{6pt}
\adjustbox{width=0.97\textwidth}{
\begin{tabular}{l| 
 c c | c c |
 c c |c c }
\toprule[0.1em]
\textbf{Shots}
& \multicolumn{2}{c|}{\rotatebox{70}{\textbf{GPT-5}}}
& \multicolumn{2}{c|}{\rotatebox{70}{\textbf{GPT-4o}}}
& \multicolumn{2}{c|}{\rotatebox{70}{\textbf{Qwen3-VL-4B-Instruct}}}
& \multicolumn{2}{c}{\rotatebox{70}{\textbf{Qwen3-VL-30B-Instruct}}} \\
\cmidrule(lr){2-3}
\cmidrule(lr){4-5}
\cmidrule(lr){6-7}
\cmidrule(lr){8-9}
&
\textbf{0}
& \textbf{2}
& \textbf{0}
& \textbf{2}
& \textbf{0}
& \textbf{2}
& \textbf{0}
& \textbf{2} \\
\toprule[0.1em]

Edges                                & 27.17                         & \cellcolor[HTML]{EA9999}29.63          & 23.91                         & 25.03                         & 23.30                         & 25.90          & 21.19          & 24.90                         \\
Location                             & 23.71                         & \cellcolor[HTML]{EA9999}26.75          & 20.62                         & 23.97                         & 21.70                         & 23.90          & 20.00          & 23.90                         \\
Rotation                             & 35.42                         & \cellcolor[HTML]{EA9999}36.37          & 26.04                         & 28.93                         & 24.20                         & 26.40          & 26.16          & 24.40                         \\
Brightness                           & 25.26                         & 26.30                                  & \cellcolor[HTML]{EA9999}27.27 & 25.00                         & 24.90                         & 25.30          & 23.34          & 24.30                         \\
Blur                                 & \cellcolor[HTML]{EA9999}31.18 & 30.90                                  & 25.26                         & 27.35                         & 22.80                         & 24.30          & 21.73          & 23.30                         \\
Global\_Occ\_50                      & 42.86                         & 51.60                                  & 20.88                         & \cellcolor[HTML]{EA9999}54.18 & 22.80                         & 23.80          & 26.20          & 25.80                         \\
Global\_Occ\_80                      & 31.10                         & \cellcolor[HTML]{EA9999}35.28          & 32.61                         & 34.94                         & 22.40                         & 23.50          & 26.60          & 25.50                         \\
\midrule
\hspace{6mm}\textbf{Level 1 Average}                     & \textbf{31.11}                & \cellcolor[HTML]{EA9999}\textbf{33.83} & \textbf{23.86}                & \textbf{31.34}                & \textbf{23.16}                & \textbf{24.73} & \textbf{23.60} & \textbf{24.59}             \\
\midrule
Orientation Uniform                  & 22.22                         & \cellcolor[HTML]{EA9999}28.17          & 25.25                         & 25.33                         & 25.11                         & 25.98          & 26.80          & 28.11                         \\
Color Uniform                        & 96.00                         & \cellcolor[HTML]{EA9999}97.30          & 21.00                         & 32.40                         & 20.80                         & 21.40          & 26.18          & 26.20                         \\
Count Uniform                        & \cellcolor[HTML]{EA9999}61.00 & 57.92                                  & 25.00                         & 27.40                         & 21.00                         & 23.60          & 27.60          & 31.20                         \\
3-different Orientation              & 34.69                         & \cellcolor[HTML]{EA9999}35.32          & 24.00                         & 26.80                         & 22.60                         & 23.20          & 28.40          & 28.60                         \\
3-different Color                    & 88.00                         & \cellcolor[HTML]{EA9999}91.40          & 26.00                         & 27.00                         & 25.60                         & 28.65          & 29.20          & 33.60                         \\
3-different Count                    & 52.00                         & \cellcolor[HTML]{EA9999}56.80          & 29.00                         & 24.20                         & 26.00                         & 25.40          & 30.20          & 31.60                         \\
1-different \& 2-similar Orientation & \cellcolor[HTML]{EA9999}31.00 & 30.18                                  & 28.00                         & 28.57                         & 25.10                         & 25.21          & 25.70          & 26.51                         \\
1-different \& 2-similar Color       & 51.00                         & \cellcolor[HTML]{EA9999}55.80          & 27.00                         & 29.00                         & 25.20                         & 24.20          & 29.00          & 28.20                         \\
1-different \& 2-similar Count       & 24.00                         & 26.85                                  & 27.27                         & 27.05                         & 26.65                         & 27.20          & 24.80          & \cellcolor[HTML]{EA9999}27.60 \\
Count Progression                    & 50.00                         & \cellcolor[HTML]{EA9999}51.00          & 13.00                         & 24.95                         & 22.20                         & 22.60          & 29.80          & 31.20                         \\
Count Arithmetic                     & \cellcolor[HTML]{EA9999}52.00 & 49.10                                  & 22.00                         & 29.40                         & 24.38                         & 24.60          & 32.64          & 31.80                         \\
Count Min-Max                        & 29.59                         & \cellcolor[HTML]{EA9999}30.86          & 22.00                         & 23.60                         & 24.26                         & 26.00          & 28.62          & 29.80                         \\
\midrule
\hspace{6mm}\textbf{Level 2 Average}                     & \textbf{49.79}                & \cellcolor[HTML]{EA9999}\textbf{50.89} & \textbf{24.12}                & \textbf{27.14}                & \textbf{24.08}                & \textbf{24.84} & \textbf{28.25} & \textbf{29.54}
\\
\midrule
Independent Color-Object-Orientation & 34.00                         & \cellcolor[HTML]{EA9999}35.24          & 25.25                         & 23.38                         & 22.36                         & 25.07          & 23.63          & 24.20                         \\
Independent Count-Object-Color       & 34.00                         & \cellcolor[HTML]{EA9999}39.54          & 24.00                         & 25.47                         & 26.26                         & 29.65          & 28.18          & 34.24                         \\
Coupled Color-Orientation            & 24.24                         & 20.97                                  & 24.00                         & 25.13                         & \cellcolor[HTML]{EA9999}26.54 & 25.40          & 24.78          & 22.80                         \\
Coupled Color-Count                  & 30.00                         & \cellcolor[HTML]{EA9999}32.66          & 22.00                         & 24.40                         & 27.00                         & 23.20          & 25.00          & 23.80                         \\
Spiral Color-Orientation             & 28.00                         & \cellcolor[HTML]{EA9999}28.43          & 22.00                         & 26.29                         & 19.71                         & 22.57          & 20.60          & 20.87                         \\
Spiral Color-Count-Object            & 56.00                         & \cellcolor[HTML]{EA9999}57.59          & 30.00                         & 23.49                         & 19.12                         & 20.26          & 21.82          & 23.80                         \\
\midrule
\hspace{6mm}\textbf{Level 3 Average}
& \textbf{34.39}                & \cellcolor[HTML]{EA9999}\textbf{35.74} & \textbf{23.86}                & \textbf{24.69}                & \textbf{23.50}                & \textbf{24.36} & \textbf{24.00} & \textbf{24.95}                \\
\midrule
\textbf{Average (All Levels)}        & \textbf{40.58}                & \cellcolor[HTML]{EA9999}\textbf{42.22} & \textbf{24.53}                & \textbf{27.73}                & \textbf{23.68}                & \textbf{24.69} & \textbf{25.93} & \textbf{27.05} \\

\bottomrule[0.1em]
\end{tabular}}
\caption{Accuracy across VisRes benchmark levels and subtasks under fewshots.}
\label{few_shots}
\end{table*}

\begin{table*}[!htbp]
\centering
\setlength{\tabcolsep}{6pt}
\adjustbox{width=0.97\textwidth}{
\begin{tabular}{l| 
 c c | c c |
 c c |c c }
\toprule[0.1em]
\textbf{Reasoning}
& \multicolumn{2}{c|}{\rotatebox{70}{\textbf{GPT-5}}}
& \multicolumn{2}{c|}{\rotatebox{70}{\textbf{Qwen3-VL-4B}}}
& \multicolumn{2}{c|}{\rotatebox{70}{\textbf{Qwen3-VL-30B}}}
& \multicolumn{2}{c}{\rotatebox{70}{\textbf{Mimo}}} \\
\cmidrule(lr){2-3}
\cmidrule(lr){4-5}
\cmidrule(lr){6-7}
\cmidrule(lr){8-9}
&
\textbf{Low}
& \textbf{High}
& \textbf{\xmark }
& \textbf{\cmark}
& \textbf{\xmark }
& \textbf{\cmark}
& \textbf{\xmark }
& \textbf{\cmark} \\
\toprule[0.1em]
Edges                                & \cellcolor[HTML]{EA9999}29.82          & 27.17                                  & 23.30          & 16.67                         & 21.19                         & 25.00                         & 22.30          & 22.30          \\
Location                             & \cellcolor[HTML]{EA9999}26.30          & 23.71                                  & 21.70          & 23.16                         & 20.00                         & 22.40                         & 20.50          & 25.77          \\
Rotation                             & 34.86                                  & 35.42                                  & 24.20          & \cellcolor[HTML]{EA9999}37.50 & 26.16                         & 36.05                         & 21.50          & 29.17          \\
Brightness                           & 24.61                                  & 25.26                                  & 24.90          & \cellcolor[HTML]{EA9999}31.52 & 23.34                         & 29.47                         & 22.50          & 27.37          \\
Blur                                 & 30.63                                  & \cellcolor[HTML]{EA9999}31.18          & 22.80          & 24.73                         & 21.73                         & 24.28                         & 21.60          & 26.32          \\
Global\_Occ\_50                      & 42.25                                  & 42.86                                  & 22.80          & 37.50                         & 26.20                         & \cellcolor[HTML]{EA9999}47.25 & 33.08          & 48.35          \\
Global\_Occ\_80                      & 31.52                                  & 31.10                                  & 22.40          & 25.88                         & 26.60                         & \cellcolor[HTML]{EA9999}35.87 & 25.90          & 30.43          \\
\midrule
\hspace{6mm}\textbf{Level 1 Average}                     & \cellcolor[HTML]{EA9999}\textbf{31.43} & \textbf{31.11}                         & \textbf{23.16} & \textbf{28.17}                & \textbf{23.60}                & \textbf{31.20}                & \textbf{23.91} & \textbf{29.22} \\
\midrule
Orientation Uniform                  & 21.76                                  & 22.22                                  & 25.11          & 26.00                         & \cellcolor[HTML]{EA9999}26.80 & 23.00                         & 24.71          & 19.19          \\

Color Uniform                        & 90.19                                  & \cellcolor[HTML]{EA9999}96.00          & 20.80          & 66.20                         & 26.18                         & 88.00                         & 27.25          & 78.95          \\
Count Uniform                        & 57.28                                  & \cellcolor[HTML]{EA9999}61.00          & 21.00          & 40.82                         & 27.60                         & 59.00                         & 26.48          & 52.75          \\
3-different Orientation              & 32.24                                  & \cellcolor[HTML]{EA9999}34.69          & 22.60          & 21.47                         & 28.40                         & 17.00                         & 25.89          & 24.24          \\
3-different Color                    & 87.84                                  & \cellcolor[HTML]{EA9999}88.00          & 25.60          & 45.92                         & 29.20                         & 65.00                         & 27.45          & 46.46          \\
3-different Count                    & 46.01                                  & \cellcolor[HTML]{EA9999}52.00          & 26.00          & 39.87                         & 30.20                         & 44.00                         & 25.61          & 34.41          \\
1-different \& 2-similar Orientation & 31.04                                  & 31.00                                  & 25.10          & 29.32                         & 25.70                         & \cellcolor[HTML]{EA9999}36.00 & 24.64          & 25.00          \\
1-different \& 2-similar Color       & 48.72                                  & 51.00                                  & 25.20          & 36.10                         & 29.00                         & \cellcolor[HTML]{EA9999}59.00 & 27.32          & 55.21          \\
1-different \& 2-similar Count       & 22.79                                  & 24.00                                  & 26.65          & 31.34                         & 24.80                         & \cellcolor[HTML]{EA9999}36.00 & 29.20          & 35.71          \\
Count Progression                    & 49.52                                  & \cellcolor[HTML]{EA9999}50.00          & 22.20          & 37.20                         & 29.80                         & 48.00                         & 28.29          & 36.96          \\
Count Arithmetic                     & 50.55                                  & \cellcolor[HTML]{EA9999}52.00          & 24.38          & 43.20                         & 32.64                         & 49.00                         & 28.06          & 33.33          \\
Count Min-Max                        & 26.14                                  & 29.59                                  & 24.26          & 28.74                         & 28.62                         & \cellcolor[HTML]{EA9999}37.00 & 25.29          & 29.35          \\
\midrule
\hspace{6mm}\textbf{Level 2 Average}                     & \textbf{47.01}                         & \cellcolor[HTML]{EA9999}\textbf{49.79} & \textbf{24.08} & \textbf{37.18}                & \textbf{28.25}                & \textbf{46.75}                & \textbf{26.68} & \textbf{39.15} \\
\midrule
Independent Color-Object-Orientation & \cellcolor[HTML]{EA9999}36.20          & 34.00                                  & 22.36          & 27.39                         & 23.63                         & 32.60                         & 23.95          & 19.00          \\
Independent Count-Object-Color       & 31.97                                  & 34.00                                  & 26.26          & 29.45                         & 28.18                         & \cellcolor[HTML]{EA9999}36.34 & 26.99          & 29.00          \\
Coupled Color-Orientation            & 23.39                                  & 24.24                                  & 26.54          & 26.13                         & 24.78                         & \cellcolor[HTML]{EA9999}29.43 & 25.67          & 20.00          \\
Coupled Color-Count                  & 27.98                                  & 30.00                                  & 27.00          & 27.46                         & 25.00                         & \cellcolor[HTML]{EA9999}33.33 & 23.25          & 28.00          \\
Spiral Color-Orientation             & 26.62                                  & \cellcolor[HTML]{EA9999}28.00          & 19.71          & 17.31                         & 20.60                         & 20.45                         & 24.23          & 22.00          \\
Spiral Color-Count-Object            & 51.17                                  & \cellcolor[HTML]{EA9999}56.00          & 19.12          & 28.63                         & 21.82                         & 36.00                         & 27.30          & 33.00          \\
\midrule
\hspace{6mm}\textbf{Level 3 Average}                     & \textbf{32.89}                         & \cellcolor[HTML]{EA9999}\textbf{34.39} & \textbf{23.50} & \textbf{26.31}                & \textbf{24.00}                & \textbf{31.36}                & \textbf{25.23} & \textbf{25.17} \\
\midrule
\textbf{Average (All Levels)}        & \textbf{39.26}                         & \cellcolor[HTML]{EA9999}\textbf{40.58} & \textbf{23.68} & \textbf{31.98}                & \textbf{25.93}                & \textbf{38.78}                & \textbf{25.56} & \textbf{33.29}\\
\bottomrule[0.1em]
\end{tabular}}
\caption{Accuracy across VisRes benchmark levels and subtasks under reasoning efforts.}
\label{reasoning}
\end{table*}
\end{document}